\documentclass[11pt]{article}

\usepackage{acl}
\usepackage{times}
\usepackage{latexsym}
\usepackage[T1]{fontenc}
\usepackage{multirow}

\usepackage{gmnlp}

\usepackage[scaled=0.8]{beramono}

\usepackage[T1]{fontenc}

\DeclareTextFontCommand{\textttenglish}{\ttenglishfamily}

\usepackage{microtype}

\newif\iffinal
\finaltrue

\title{Dataset Geography: Mapping Language Data to Language Users}

\author{Fahim Faisal, Yinkai Wang, Antonios Anastasopoulos \\
  Department of Computer Science, George Mason University, USA \\
  \texttt{\{ffaisal, ywang88, antonis\}@gmu.edu} }
\begin{document}

\maketitle
\begin{abstract}
As language technologies become more ubiquitous, there are increasing efforts towards expanding the language diversity and coverage of natural language processing (NLP) systems. Arguably, the most important factor influencing the quality of modern NLP systems is data availability. 
In this work, we study the geographical representativeness of NLP datasets, aiming to quantify if and by how much do NLP datasets match the expected needs of the language speakers. In doing so, we use entity recognition and linking systems, presenting an approach for good-enough entity linking without entity recognition first.  
Last, we explore some geographical and economic factors that may explain the observed dataset distributions.\footnote{Code and data are publicly available: \url{https://github.com/ffaisal93/dataset_geography}. Additional visualizations are available in the project page: \url{https://nlp.cs.gmu.edu/project/datasetmaps/}.}
\end{abstract}

\section{Introduction}

The lack of linguistic, typological, and geographical diversity in NLP research, authorship, and publications is by now widely acknowledged and documented~\cite{caines,ponti-etal-2019-modeling,bender2011achieving,adelani2021masakhaner}. Nevertheless, the advent of massively multilingual models presents opportunity and hope for the millions of speakers of under-represented languages that are currently under-served by language technologies.

Broadening up the NLP community's research efforts and scaling from a handful up to the almost 7000 languages of the world is no easy feat. In order for this effort to be efficient and successful, the community needs some necessary foundations to build upon. In seminal work, \citet{joshi-etal-2020-state} provide a clear overview of where we currently stand with respect to data availability for the world's languages and relate them to the languages' representation in NLP conferences. \citet{choudhury2021linguistically} study how linguistically fair are multilingual language models, and provide a nuanced framework for evaluating multilingual models based on the principles of fairness in economics and social choice theory. Last, \citet{blasi-etal-2021-systematic} provide a framework for relating NLP systems' performance on benchmark datasets to their downstream utility for users at a global scale, which can provide insights into development priorities; they also discuss academic incentives and socioeconomic factors that correlate with the current status of systematic cross-lingual inequalities they observe in language technologies performance.

\begin{figure}[t]
    \centering
    \begin{tabular}{c}
        \includegraphics[width=.4\textwidth]{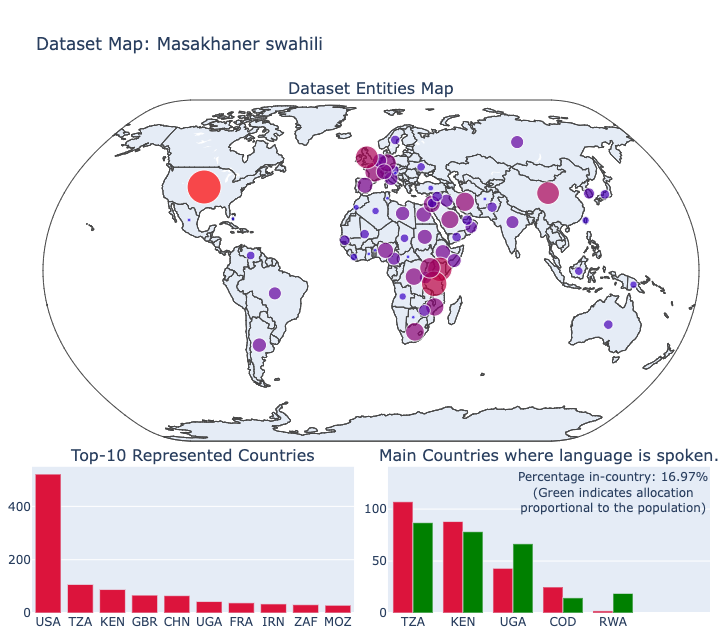} \\
    \end{tabular}
    \caption{Example of the dataset map our method produces for the Swahili section of MasakhaNER. The dataset is only somewhat representative of Swahili speakers, with only about 17\% of entity mentions related to Tanzania, Kenya, Uganda, DR.\ Congo, or Rwanda and neighboring countries, with the USA and western Europe over-represented.}
    \label{fig:ex_maps}
    \vspace{-1em}
\end{figure}

These works provide insights into current data availability and estimated utility that are paramount for making progress, as well as an evaluation framework for future work. However, there is one missing building block necessary for \textit{real} progress: a way to estimate how representative of the underlying language speakers is the content of our datasets. Any evaluation framework and any utility estimates we build can only be trustworthy as long as the evaluation data are representative. \citet{gebru2021datasheets} and \citet{bender2018data} recognize the importance of this information, including them in their proposed guidelines for ``datasheets'' and ``data statements'' respectively; but most datasets unfortunately lack such meta-information. To the best of our knowledge, MaRVL~\cite{liu-etal-2021-visually} is the only dataset that is culturally-aware \textit{by design} in terms of its content.\footnote{Datasets designed to capture dialectal variations, e.g., SD-QA~\cite{faisal-etal-2021-sd-qa}, are culturally-aware in terms of annotator selection, but there is no guarantee that their content is also culturally-relevant for the language speakers.}

We propose a method to estimate a dataset's cultural representativeness by mapping it onto the physical space that language speakers occupy, producing visualizations such as Figure~\ref{fig:ex_maps}.
Our contributions are summarized below:
\begin{itemize}[leftmargin=*]
    \item We present a method to map NLP datasets unto geographical areas (in our case, countries) and use it to evaluate how well the data represent the underlying users of the language. We perform an analysis of the socio-economic correlates of the dataset maps we create. We find that dataset representativeness largely correlates with economic measures (GDP), with geographical proximity and population being secondary.
    \item We test a simple strategy for performing entity linking by-passing the need for named entity recognition. We evaluate its efficacy on 19 languages, showing that we can get within up to 85\% of a NER-informed harder-to-obtain model. We also show that encouragingly, using either model largely leads to similar dataset maps.
\end{itemize}

\section{Mapping Datasets to Countries}
\paragraph{Assumptions} This work makes two assumptions: that (a) data locality matters, i.e., speakers of a language are more likely to talk about or refer to local news, events, entities, etc as opposed to ones from a different side of the world, and (b) that we can capture this locality by only focusing on entities. \citet{kumar-etal-2019-topics} discuss these \textit{topical correlations} that are present in datasets,\footnote{See \S2 of their paper.} noting that they exist and that L1 language identification models tend to pick up on them, i.e. if a text mentions Finland, a L1 langid model is probably going to predict that the speaker is Finnish, because $p(\mathtt{Finland}\vert\mathtt{L1=Finnish})$ is generally high. In that work \citet{kumar-etal-2019-topics} make explicit effort to avoid learning such correlations because they are interested in building models for $p(\mathtt{L1}\vert\mathtt{text})$ (i.e. $p(\mathtt{L1=Finnish}\vert\mathtt{Finland})$) that are not confounded by the reverse conditional. The mere fact they need to do this, though, confirms that real-world text has such topical confounds.

As for our second assumption that we can capture these topical correlations by only looking at entities, one need only to take a look at Table~2 of~\citet{kumar-etal-2019-topics}, which lists the top topical confounding words based on log-odds scores for each L1 language in their dataset: all lists include either entities related to a country where that language is spoken (e.g. `Merkel', the name of a former chancellor, for German) or topical adjectives (e.g. `romanian' for Romanian).


\paragraph{Approach} For a given dataset, our method follows a simple recipe:
\begin{enumerate}
    \item Identify named entities present in the dataset. 
    \item Perform entity linking to wikidata IDs.
    \item Use Wikidata to link entities to countries.
\end{enumerate}
We discuss each step below.

\paragraph{Entity Recognition Step} 
Standard entity linking is treated as the sequence of two main tasks: entity recognition and entity disambiguation. One approach is to first process the text to extract entities and then disambiguate these entities to the correct entries of a given knowledge base (eg. Wikipedia). This approach relies on NER model quality.

However, to perform analysis on several datasets spanning several low-resource languages, one needs good-quality NER models in all these languages. 
The interested reader will find a discussion on the cross-lingual consistency of NER models in Appendix~\ref{sec:consistency}.\footnote{Discussion summary: state-of-the-art NER models are \textit{not} cross-lingually consistent, i.e. they do not produce the same entity labels when presented with translations of the same sentence. We recommend using parallel data as part of the evaluation sets in multiple languages to measure this important aspect of models' performance.} As we show in Section~\S\ref{sec:bypassing}, we can bypass this NER step if we tolerate a small penalty in accuracy.

\paragraph{Entity Linking Step} In this step we map named entities to their respective Wikidata IDs. We further discuss this step in Section~\S\ref{sec:bypassing}.

\paragraph{From Entities to Countries} 
We produce maps to visualize the geographical coverage of the datasets we study, discussing their properties and our findings in Section~\S\ref{sec:maps}. 

To link entities to countries,\footnote{A single entity can be associated with a set of more than one countries.} we rely on Wikidata entries, depending on the type of entity:
\begin{itemize}[noitemsep,nolistsep,leftmargin=*]
    \item for persons, we log their places of birth (P19) and death (P20), and country of citizenship~(P27);
    \item for locations, we search for their associated country (P17); and
    \item for organizations, we use the links of the `located\_at' (P276) and `headquartered\_at' (P159) relations. 
\end{itemize}
Since places of birth/death and headquarters are not necessarily at the country level, we perform a second step of associating these locations with countries. In cases where the result does not correspond to a modern-day country (as can often be the case with historical figures), we do not make any attempts to link it to any modern day countries, excluding them from the analysis. 

\iffinal
For example, the entry for Nicolaus Copernicus (Q619) lists him as born in Toru\'{n} (Q47554) which is then mapped to Poland; as having died in Frombork (Q497115) that also maps to Poland; and as a citizen of the Kingdom of Poland (Q1649871) which is not mapped to any modern-day country; so he is only linked to Poland. Albert Einstein is similarly mapped to both Germany and the United States, due to his places of birth (Ulm) and death (Princeton).
\fi

\section{Dataset-Country Maps}
\label{sec:maps}
\begin{figure*}[t]
    \centering
    \begin{tabular}{cc}
    \multicolumn{1}{c}{\textbf{Natural Questions}} & \textbf{MLQA}\\
        \includegraphics[width=.48\textwidth]{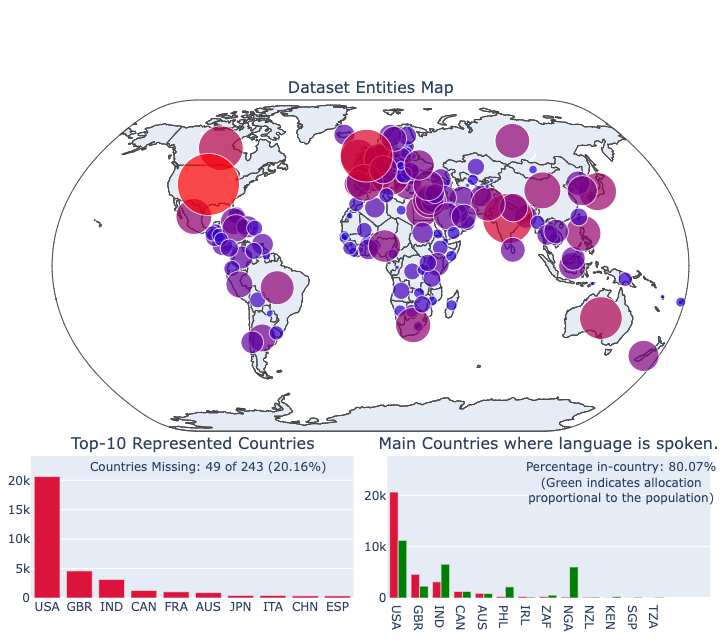} & 
        \includegraphics[width=0.48\textwidth]{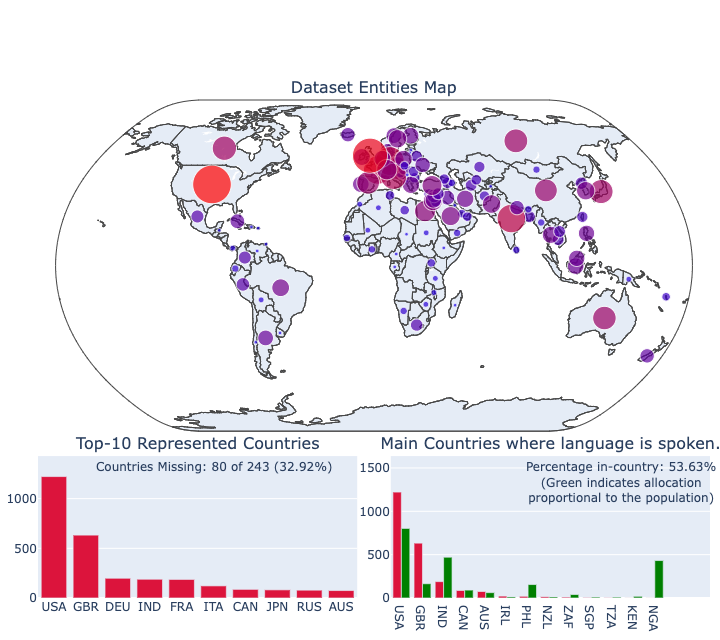}\\
    \multicolumn{1}{c}{\textbf{SQuAD}} & \textbf{TyDi-QA (English)}\\
        \includegraphics[width=.48\textwidth]{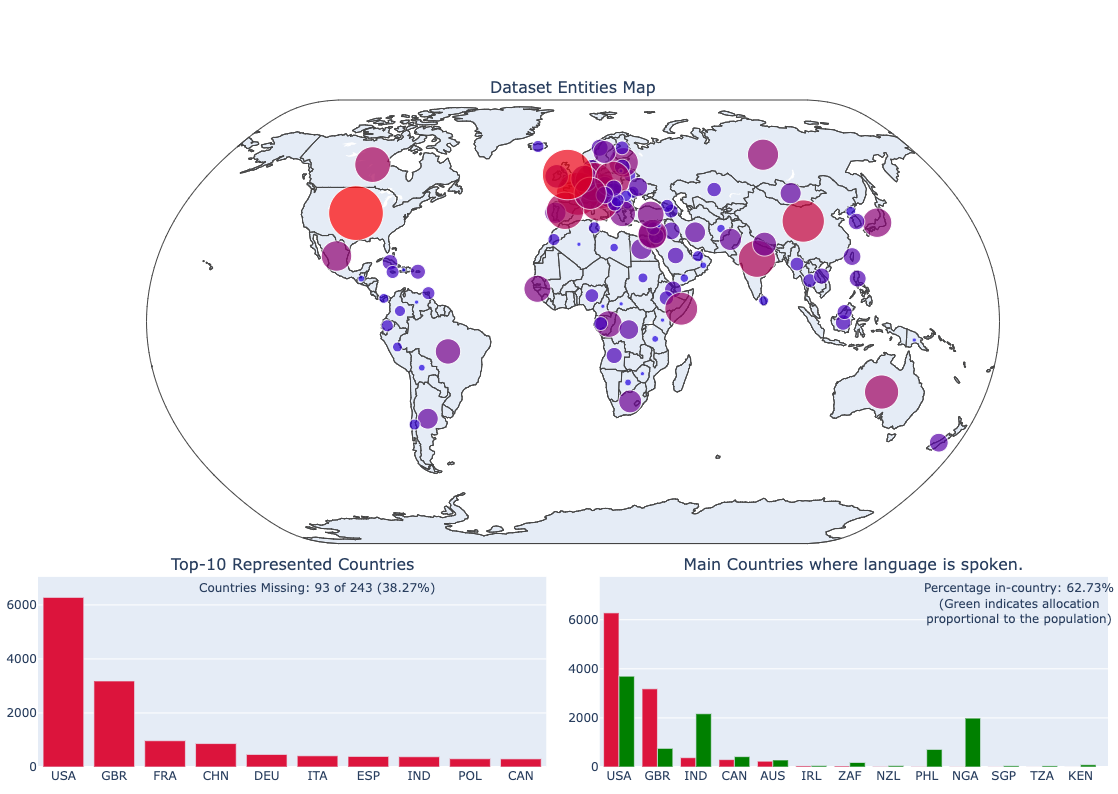} &  \includegraphics[width=.4\textwidth]{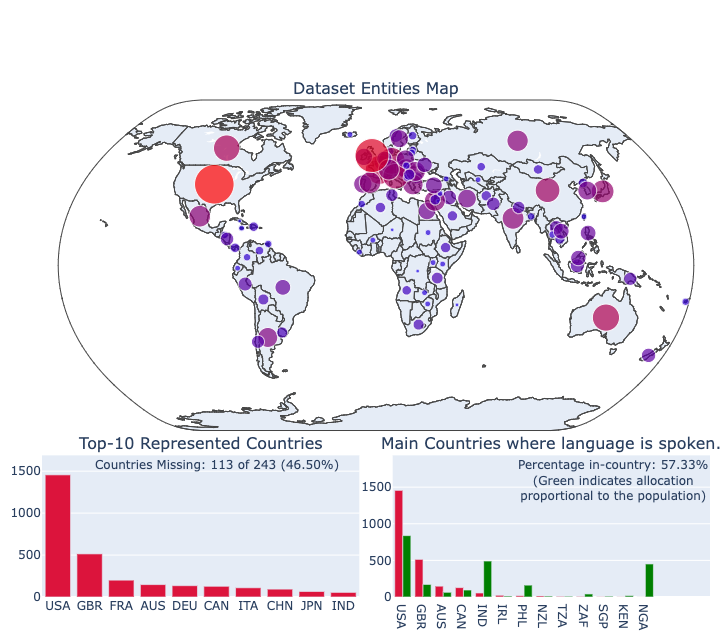} \\
    \end{tabular}
    \caption{Visualizing the datasets' geography allows easy comparisons of their representativeness (best viewed in color and zoomed-in). NQ is the most representative of English speakers, with in-country percentage (higher is better) of 80\% (SQuAD: 63\%; TyDi-QA: 57\%; MLQA: 53\%) and less countries left unrepresented (lower is better; NQ: 49; MLQA: 80; SQuAD: 93; TyDi-QA: 113).}
    \label{fig:maps}
    \vspace{-1em}
\end{figure*}

Before delving into our case studies, we first list a set of statistics of interest that one could extract from our produced dataset-country maps, in order to gauge a dataset's representativeness.

\paragraph{Representativeness Measures} We will avoid providing a single metric, largely because the ideal metric to use will be very dataset-specific and related to the goals of the creators of the dataset and the socioeconomic correlates they are interested in (see discussion in Section~\S\ref{sec:correlates}).

As a first straightforward representativeness measure, we will compute the \textbf{percentage of entities associated with countries where the language is largely spoken}. For example, according to Ethnologue~\cite{ethnologue}, most Swahili speakers\footnote{In the case of Swahili they are often second-language speakers.} reside in Tanzania, Kenya, Uganda, DR. Congo, and Rwanda. For a Swahili dataset, then, we compute the percentage of all entities associated with this set of countries (``\textit{in-country}'').

Notions of equity or fairness across countries could be measured by various fairness metrics, given the distribution of entities over countries in a dataset: from simply computing the standard deviation of the observations,\footnote{Or approximations thereof such as the max-min of the observations, as used by~\cite{debnath-etal-2021-towards}.} to treating countries as a population and computing fairness indices like the popular Gini index~\cite{gini1912variabilita,gastwirth1972estimation} or the indices proposed by~\citet{speicher2018unified}. We will opt for a simpler, much more interpretable measure, \textbf{the number of countries not represented in the dataset} i.e. countries with associated entity count below a given threshold (we use~zero for simplicity but higher values would also be reasonable for large datasets).

Last, especially for languages with significant amounts of speakers in more than one country, it is important to go deeper and measure the representativeness of this \textit{in-country} portion. For a simple example, an English dataset with entities only from the UK is probably not representative of Nigerian or Jamaican English speakers. Hence, we will create two distributions over the countries where the language is largely spoken: the distribution of speaker populations (as available from Ethnologue and other public data), and the distribution of entities observed in the dataset. Discrepancies between these two distributions will reveal potential issues. While one could easily compute some measure of distance between the two distributions (e.g. the Bhattacharyya coefficient \cite{Bhattacharyya}), in this work we will rely on the interpretable advantages of the visualizations. Measures of fairness could be computed for this portion of the dataset, similarly as discussed above.

In the example dataset of the Swahili portion of MasakhaNER in Figure~\ref{fig:ex_maps}, the utility of our method is apparent. Through the visualization, a researcher can quickly confirm that the dataset seems to not reflect the users of the language to a large extent: only about 17\% of the entities indeed correspond 
to Tanzania, Kenya, Uganda, DR. Congo, or Rwanda (where Swahili and its varieties are treated as a lingua franca, at least in portions of these countries). Wealthy or populous countries like USA, France, and China, are  well-represented,\footnote{over-represented?} as one would expect, while 156 countries and territories have no representation. At the same time, the visualization allows a researcher to identify gaps: beyond the neighboring African countries and perhaps the Middle East, north-west African countries as well as central America or central/south-east Asia are clearly under-represented in this dataset. Between the main Swahili-speaking countries, Tanzania, Kenya, and Uganda are well-represented (DR Congo and Rwanda less so, but they have less Swahili speakers), with the former two perhaps slightly over-represented and the latter (as well as Rwanda) being under-represented relative to the speakers population, c.f. red (dataset entities) and green (proportional to population) bars in Figure~\ref{fig:ex_maps}.

\subsection{Datasets and Settings} 
We apply the process described above on several datasets, chosen mostly for their language and typological diversity. Our process is not dataset- or language-dependent,\footnote{Although it does rely on a decent quality entity linker which we lack for most languages. See discussion in \S\ref{sec:bypassing}.} and could easily be applied on any NL dataset.
We briefly describe the datasets we include in our study below, with detailed statistics in Appendix~\ref{app:d_stat}. 

\paragraph{NER Datasets} We study the WikiANN dataset~\cite{pan-etal-2017-cross} that is commonly used in the evaluation of multilingual models. 
We additionally study the MasakhaNER dataset~\cite{adelani2021masakhaner}, which was created through participatory design~\cite{nekoto2020participatory} in order to focus on African languages. Since these datasets are already annotated with named entities, we only need to perform entity linking. 

\paragraph{Question Answering} We study four question answering datasets (focusing on the questions rather than contexts), namely SQuAD~\cite{rajpurkar2016squad},  MLQA~\cite{lewis2020mlqa}, TyDi-QA~\cite{clark2020tydi}, and Natural Questions~\cite[NQ;]{nq}, which have unique characteristics that lend themselves to interesting comparisons. SQuAD is a large English-only dataset (although it has been translated through efforts like XQuAD~\cite{artetxe-etal-2020-cross}). MLQA is a $n$-way parallel multilingual dataset covering 7 languages, created by translating an English dataset. TyDi-QA is another multilingual dataset covering 11 languages, but each language portion is derived separately, without translation involved. Last, NQ is an English QA dataset created based on real-world queries on the Google search engine for which annotators found relevant Wikipedia context, unlike the other datasets that were created by annotators forming questions \textit{given} a context.

\iffinal
\paragraph{Additional Datasets} While not further discussed in this paper, additional visualizations for more datasets (e.g. for the X-FACTR benchmark~\cite{jiang-etal-2020-x}, and several machine translation benchmarks) are available in the project's webpage: \url{https://nlp.cs.gmu.edu/project/datasetmaps/}. 
\fi

\subsection{Discussion} Beyond Figure~\ref{fig:ex_maps}, we also show example maps in Figure~\ref{fig:maps} for NQ, MLQA, SQuAD, and the English portion of TyDi-QA. We provide additional maps for all other datasets in Appendix~\ref{app:maps}. 

\paragraph{Comparing datasets} 
The comparison of MasakhaNER to the WikiANN dataset (see Appendix~\ref{app:maps}) reveals that the former is rather more localized (e.g. more than 80\% of the identified entities in the Dholuo dataset are related to Kenya) while the latter includes a smaller portion from the countries where most native speakers reside (between 10\%-20\%) and almost always also includes several entries that are very European- or western-centric. 

The effect of the participatory design~\cite{nekoto2020participatory} approach on creating the MasakhaNER dataset, where data are curated from local sources, is clear in all language portions of the dataset, with data being highly representative of the speakers. In Figures~\ref{fig:masakhaner1}--\ref{fig:masakhaner2} (App.~\ref{app:maps}) the majority of entities in the Wolof portion are from Senegal and neighboring countries (as well as France, the former colonial power of the area), and the Yoruba and Igbo ones are centered on Nigeria.

Figure~\ref{fig:maps} allows for a direct comparison of different QA datasets (also see maps for other TyDi-QA languages in Appendix~\ref{app:maps}). The first notable point has to do with NQ, which was built based on real-world English-language queries to the Google search engine. Since such queries happen all over the world, this is reflected in the dataset, which includes entities from almost all countries in the world. Two types of countries are particularly represented: ones where English is an official language (USA, UK, Australia, but also, to a lesser extent, India, Nigeria, South Africa, and the Philippines); and wealthy ones (European, Japan, China, etc). In our view, NQ is an exemplar of a representative dataset, because it not only includes representation of most countries where the language is spoken (with the sum of these entities being in their large majority in-country: 80\%) but due to its size it also includes entities from almost all countries.

\begin{table*}[t]
    \centering
    \small
    \vspace{-1em}
    \begin{tabular}{lrr|rr|rr|rr}
    \toprule
        & \multicolumn{2}{c}{TyDi-QA (11)} & \multicolumn{2}{c}{MLQA (1)} & \multicolumn{2}{c}{SQUAD (1)} & \multicolumn{2}{c}{NaturalQ. (1)} \\
        \textbf{Factors $\phi$} & \textbf{Expl. Var.} & \textbf{MAE}  & \textbf{Expl. Var.} & \textbf{MAE}   & \textbf{Expl. Var.} & \textbf{MAE} & \textbf{Expl. Var.} & \textbf{MAE}  \\
    \midrule
         \texttt{pop} & 0.272 & 0.431  & 0.317  & 0.401  & 0.277 & 1.230 & 0.395 & 1.18\\ 
         \texttt{gdp} & 0.507 & 0.349  & 0.561 & 0.332  & 0.516 & 1.023 & 0.535 & 1.069\\ 
         \texttt{gdppc} & 0.176 & 0.458  & 0.182 & 0.458 & 0.127 & 1.345 & 0.144 & 1.463 \\ 
         \texttt{land} & 0.107 & 0.504 & 0.166 & 0.469 & 0.142 & 1.380 & 0.152 & 1.459\\ 
         \texttt{geo} & 0.075 & 0.499  & 0.040 & 0.495 & 0.062 & 1.393 & 0.030 & 1.561\\ 
    \midrule
        \texttt{geo+gdp} & 0.550 & 0.333  & \textbf{0.579} & 0.321 & \textbf{0.552} & 0.932 & 0.550 & 1.054\\ 
        \texttt{pop+gdp+geo} & 0.532 & 0.337  & 0.548 & 0.326 & 0.534 & 0.940 & 0.550 & 1.005 \\
        \texttt{pop+gdp+gdppc+geo} & \textbf{0.555} & \textbf{0.321} & 0.576 & \textbf{0.310} & 0.531 & \textbf{0.918} & \textbf{0.570} & \textbf{0.973} \\
    \midrule
        \texttt{all 5 factors} & 0.538 & 0.325 & 0.566 & 0.312 & 0.524 & 0.924  & 0.561 & 0.981\\ 
    \bottomrule
    \end{tabular}
    \vspace{-1em}
    \caption{Empirical comparison of factors on QA datasets, averaging over their respective languages (number in parentheses). We report the five-fold cross-validation explained variance and mean absolute error of a linear model.}
    \label{tab:factors}
    \vspace{-1em}
\end{table*}
SQuAD also has a large percentage in-country (63\%) but it is less representative of different Englishes than NQ. India, for instance, is relatively under-represented in all datasets; in SQuAD it ranks 7\textsuperscript{th}, but it ranks 3\textsuperscript{rd} in NQ (see red bars in bottom left of figures).
On the other hand, the geographical representativeness of both MLQA and TyDi-QA (their English portion) is lacking. Since these datasets rely on Wikipedia articles for their creation, and Wikipedia has a significant western-country bias~\cite{greenstein2012wikipedia,hube2018detecting}, most entities come from Europe, the US, and the Middle East. All these datasets under-represent English speakers from English-speaking countries of the Global South like Kenya, South Africa, or Nigeria, since there are practically almost no entities from these countries. MLQA further under-represents the speakers of all other languages it includes beyond English, since all data are translations of the English one. Contrast this to TyDi-QA and its visualized Swahili portion which, even though still quite western-centric, does have a higher representation from countries where Swahili is spoken than the TyDi-QA English portion.

This discussion brings forth the importance of being cautious with claims regarding systems' utility, when evaluated on these datasets. One could argue that a QA system that is evaluated on NQ does indeed give a good estimation of real-world utility; a system evaluated on TyDi-QA gives a distorted notion of utility (biased towards western-based speakers and against speakers from the Global South); a system evaluated on MLQA will give an estimation as good as one evaluated on TyDi-QA, but only on the English portion. We clarify that this does not diminish the utility of the datasets themselves as tools for comparing models and making progress in NLP: MLQA is extremely useful for comparing models across languages \textit{on the exact same data}, thus facilitating easy comparisons of the cross-lingual abilities of QA systems, without the need for approximations or additional statistical tests. But we argue that MLQA should not be used to asses the potential utility of QA systems for German or Telugu speakers.

Similar observations can be made about comparing two similar projects that aim at testing the memorization abilities of large language models, namely X-FACTR and multi-LAMA~\cite[mLAMA;][]{kassner-etal-2021-multilingual} -- see corresponding Figures in Appendix~\ref{app:maps}. Both of these build on top of Wikidata and the mTREx dataset. However, mLAMA translates English prompts and uses entity-relation triples mined from the English portion of Wikidata, unlike X-FACTR which uses different data for each language, mined from their respective portion of Wikidata. Both are still western-biased, since they rely on Wikipedia, but one (X-FACTR) is better at giving an indication of potential downstream utility to users.

\subsection{Socioeconomic Correlates}
\label{sec:correlates}
In this section we attempt to explain our findings from the previous section, tying them to socioeconomic factors.

\paragraph{Empirical Comparison of Factors}
We identify socioeconomic factors $\phi$ that could be used to explain the observed geographic distribution of the entities in the datasets we study. These are:
\begin{itemize}[leftmargin=*]
    \item a country's population $\phi_{\text{pop}}$
    \item a country's gross domestic product (GDP) $\phi_{\text{gdp}}$
    \item a country's GDP per capita $\phi_{\text{gdppc}}$
    \item a country's landmass $\phi_{\text{land}}$
    \item a country's geographical distance from country/ies where the language is spoken $\phi_{\text{geo}}$
\end{itemize}

The first four factors are global and fixed.
The fifth one is relative to the language of the dataset we are currently studying. For example, when we focus on the Yoruba portion of the mTREx dataset, we use Nigeria (where Yoruba is spoken) as the focal point and compute distances to all other countries. The assumption here is that a Yoruba speaker is more likely to use or be interested in entities first from their home country (Nigeria), then from its neighboring countries (Cameroon, Chad, Niger, Benin) and less likely of distant countries (e.g. Argentina, Canada, or New Zealand). Hence, we assume the probability to be inversely correlated with the country's distance. For macro-languages or ones used extensively in more than one country, we use a population-weighted combination of the factors of all relevant countries. 

To measure the effect of such factors it is common to perform a correlational analysis, where one measures Spearman's rank correlation coefficient $\rho$ between the dataset's observed geographical distribution and the factors $\phi$. It is important, though, that the factors are potentially covariate, particularly population and GDP. 
Hence, we instead compute the variance explained by a linear regression model with factors $\phi$ as input, i.e., $a\phi_{\text{pop}} + b\phi_{\text{gdp}} + c\phi_{\text{gdppc}} + d\phi_{\text{geo}} + e$ with  $a$--$e$ learned parameters, trained to predict the log of observed entity count of a country. We report explained variance and mean absolute error from five-fold cross-validation experiments to avoid overfitting.

\paragraph{Socioeconomic Correlates and Discussion} 
The results with different combination of factors for the QA datasets are listed in Table~\ref{tab:factors}.\footnote{See Appendix~\ref{app:ner_factors} for NER datasets, and Appendix~\ref{app:correlates} for a breakdown by language for all datasets.} The best \textit{single} predictor is, perhaps unsurprisingly, the GDP of the countries where the language is spoken: all datasets essentially over-represent wealthy countries (e.g. USA, China, or European ones). Note that GDP per capita is not as good a predictor, neither is landmass. A combination of geographical distance with GDP explains most of the variance we observe for all datasets, an observation that confirms the intuitions we discussed before based solely on the visualizations. Importantly, the fact that including population statistics into the model deteriorates its performance is further proof that our datasets are not representative of or proportional to the underlying populations. The only dataset that is indeed better explained by including population (and GDP per capita) is NQ, which we already argued presents an exemplar of representativeness due to its construction protocol.

\begin{figure*}[t]
    \centering
    \vspace{-1em}
    \begin{tabular}{cc}
    \includegraphics[width=.4\textwidth]{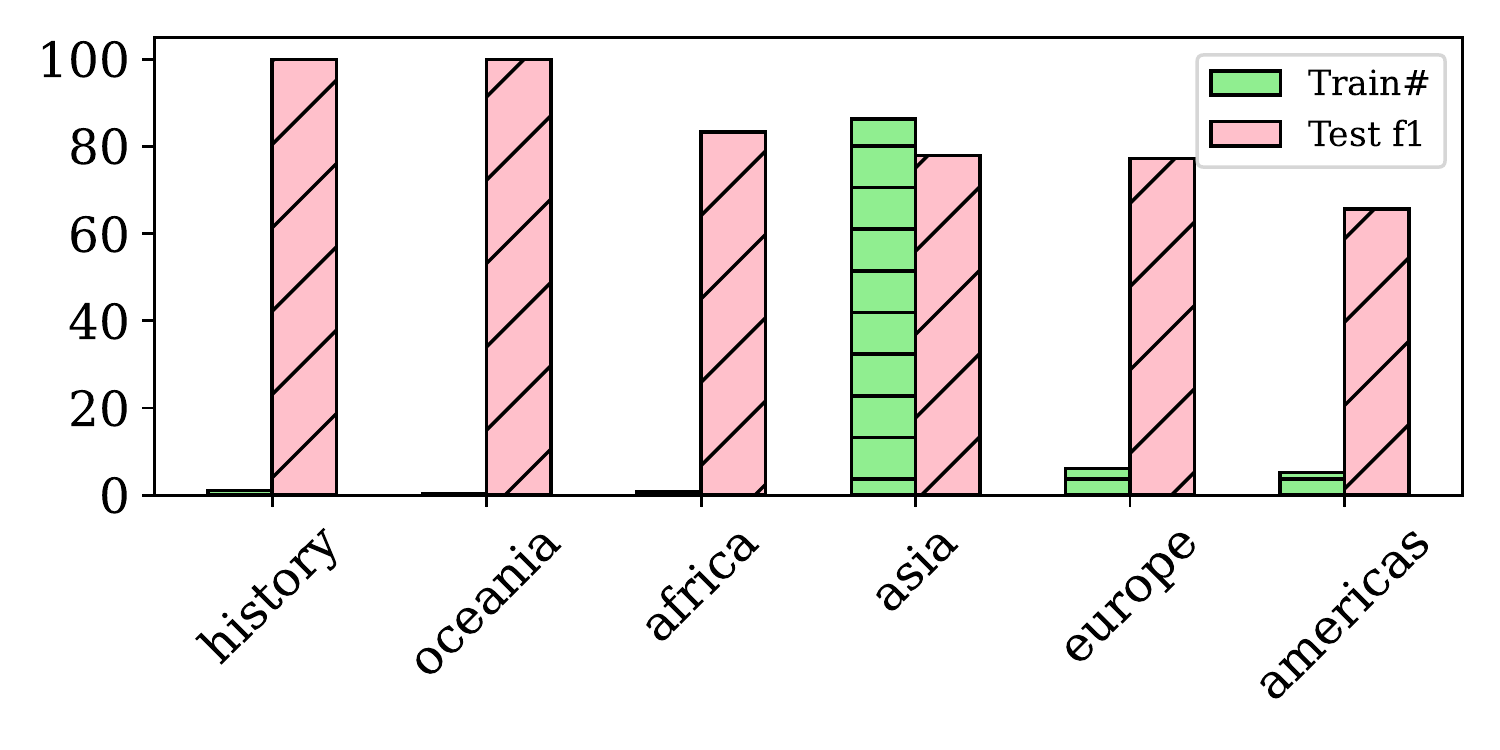} & \includegraphics[width=.4\textwidth]{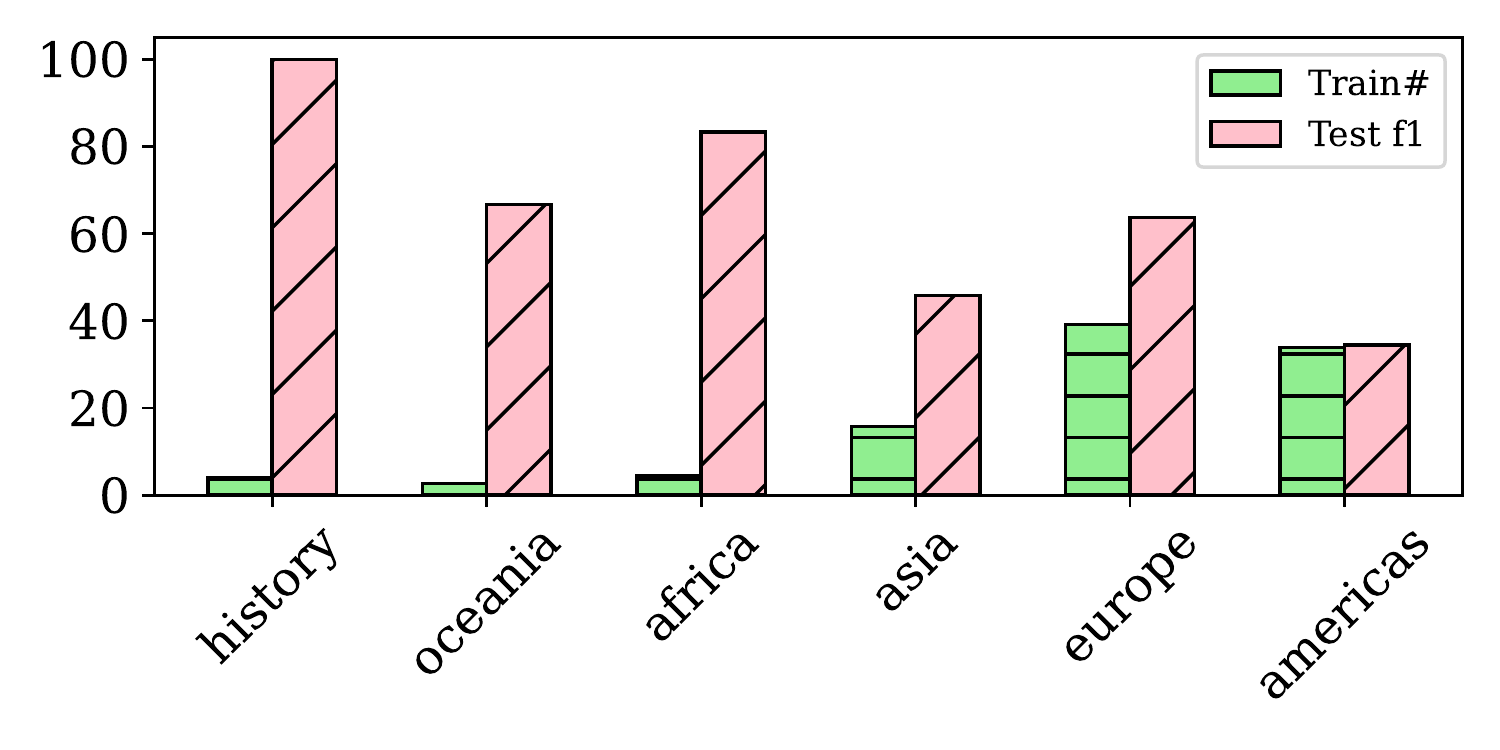}\\[-.5em]
    (a) Train: TyDiQA, Test: TyDiQA dev set  & (b) Train: (English) SQuAD, Test: TyDiQA dev set \\
    \end{tabular}
    \vspace{-1em}
    \caption{ Area-based breakdown of the performance of two models on the Telugu TyDi-QA dev set (red bars) compared with train-set distribution of these geographical areas (green bars). Model (b) is less fair than Model (a). Compare, for instance, the differences in performance between Asia and Europe of the two models.}
    \label{fig:area_comparison}
    \vspace{-1em}
\end{figure*}

\paragraph{Limitations} It is important to note that our assumptions are also limiting factors in our analyses. Mapping languages to countries is inherently lossy. It ignores, for instance, the millions of immigrants scattered throughout the world whose L1 language could be different than the dominant language(s) in the region where they reside. Another issue is that for many languages the necessary granularity level is certainly more fine than country; if a dataset does not include any entities related to the Basque country but does include a lot of entities from  Spain and France, our analysis will incorrectly deem it representative, even though the dataset could have been a lot more culturally-relevant for Basque speakers by actually including Basque-related entities.

Another limitation lies in the current state of the methods and data resources on which our approach relies. Beyond discrepancies in NER/EL across languages (addressing which is beyond the scope of this work), we suspect that Wikidata suffers from the same western-centric biases that Wikipedia is known for~\cite{greenstein2012wikipedia}. As a result, we might be underestimating the cultural representativeness of datasets in low-resource languages.

An additional hurdle, and why we avoid providing a single concrete \textit{representativeness score} or something similar, is that the ideal combination of socioeconomic factors can be subjective. It could be argued, for instance, either that geographic proximity by itself should be enough, or that it should not matter at all. Even further, other factors that we did not consider (e.g. literacy rate or web access) might influence dataset construction decisions. In any case, we share the coefficients of the NQ model, since it is the most representative dataset we studied, at least for English: $a=0.1.46$ (for $\phi_{\text{pop}}$), $b=0.87$  ($\phi_{\text{gdp}}$), $c=25.4$ ($\phi_{\text{gdppc}}$), $d=0.41$  ($\phi_{\text{geo}}$).
We believe that ideally GDP should not matter ($b\rightarrow 0$) and that a combination of speaker population and geographic proximity is ideal.\footnote{However regrettable a fact, it is undeniable that western culture and politics have world-wide effects. So their (over-)representation as a result of their high influence (and GDP) might actually reflect the true interests of people everywhere!}

\subsection{Geographical Breakdown of Models' Performance}

Beyond the analysis of the datasets themselves, we can also break down the performance of models by  geographical regions, by associating test (or dev) set samples containing entities with the geographical location of said entities. Since most test sets are rather small (a few hundred to a couple thousand instances) we have to coarsen our analysis: we map each country to a broader region (Africa, Americas, Asia, Europe, Oceania), keeping historical entities in a separate category (History).\footnote{Future work could explore a different clustering.}

We perform such a case study on TyDi-QA, comparing the performance on the TyDi-QA development sets of two models: one trained monolingually on the training set of each language of TyDi-QA (gold task), and another model trained by~\citet{debnath-etal-2021-towards} on English SQuAD and automatically generated translations in the target languages. 
Example results on Telugu shown in Figure~\ref{fig:area_comparison} reveal some notable trends.\footnote{See Table~\ref{tab:geo_f1} in Appendix \ref{app:geo_f1} for all languages.} First, training set representation (green bars in the Figures) is not a necessary condition for good test set performance (red bars). Some test set instances (e.g. with historical and African entities) receive similar test F1 score from both models. Perhaps the most interesting though, is the comparison of the Asian and European portions of the test set: the Telugu monolingual model achieves similar performance in these two subsets; but the SQuAD-trained model is almost 20 percentage points worse on the Asian subset, showing the potential unfairness of translation-based models~\cite{debnath-etal-2021-towards}. For most TyDi-QA languages (Indonesian being an exception, see Table~\ref{tab:std}) the macro-standard deviation (computed over the averages of the 6 region subsets) is larger for the SQuAD-trained model (which is, hence, less fair than models trained on TyDi-QA).

\begin{table}[t]
    \centering
    \small
    \begin{tabular}{l|cc|c}
    \toprule
         & \multicolumn{2}{c|}{{\small Stdev over the 6 regions }} & \\
        { TyDi-QA} & \multicolumn{2}{c|}{{\small of model trained on}} & \\
        { Test Set} & SQuAD & TyDi-QA & $\Delta$\\
    \midrule
        Indonesian & 17.40 &    21.52 &  -4.12 \\
        English & 13.11 &    12.66 &   0.46 \\
        Finnish & 6.33 &     5.99 &   0.3 \\
        Arabic & 19.24 &    10.08 &   9.16 \\
        Telugu & 21.83 &    12.45 &   9.38 \\
        Bengali & 36.41 &    10.21 &  26.1 \\
    \bottomrule
    \end{tabular}
    \vspace{-1em}
    \caption{Standard deviation (the lower the more fair the model) of area-based performance averages for two models. Evaluation on TyDi-QA development set.}
    \label{tab:std}
    \vspace{-1em}
\end{table}

\section{Bypassing NER for Entity Linking}
\label{sec:bypassing}

We use mGENRE~\cite{decao2021multilingual} for the task of multilingual entity linking, a sequence to sequence system that predicts entities in an auto-regressive manner. It works particularly well in a zero-shot setting as it considers 100+ target languages as latent variables to marginalize over. 

Typically, the input to mGENRE can be informed by a NER model that provides the named entity span over the source. For instance, in the Italian sentence {\small{\texttt{"[START] Einstein [END] era un fisico tedesco."}}} (\textit{Einstein was a German physicist.}) the word {\small\texttt{Einstein}} is enclosed within the entity span. mGENRE is trained to use this information to return the most relevant Wikidata entries.

Due to the plasticity of neural models and mGEBRE's auto-regressive token generation fashion, we find that by simply enclosing the whole sentence in a span also yields meaningful results. In particular, for the previously discussed Italian sentence now the input to mGENRE is {\small\texttt{"[START] Einstein era un fisico tedesco.\ [END]"}}. 

The advantage of this approach is two-fold. First, one does not need a NER component. Second, exactly because of bypassing the NER component, the EL model is now less constrained in its output; in cases where the NER component made errors, there's a higher chance that the EL model will return the correct result.

\paragraph{Experiments and Results}

\begin{figure}[t]
\centering
\vspace{-.5em}
\includegraphics[width=.48\textwidth]{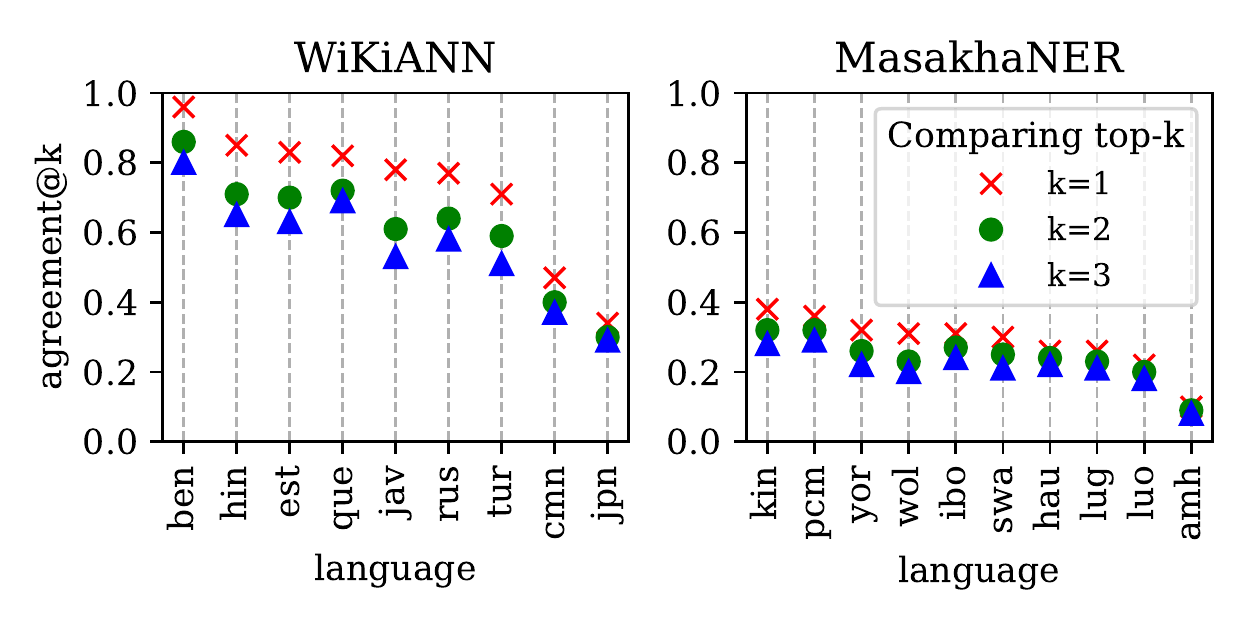}
\vspace{-2em}
\caption{For some languages a NER-Relaxed model is within 60\% of a NER-Informed model. agreement@$k$: ratio of top-$k$ agreement of the models.}
\label{fig:ner_inf}
\vspace{-1em}
\end{figure}

We conduct experiments to quantify how different a model uninformed by a 
NER model (NER-Relaxed) will perform compared to one following the typical pipeline (NER-Informed). 

Given the outputs of the two models over the same set of sentences, we will compare their average agreement@$k$, as in the size of the intersection of the outputs of the two models divided by the number of outputs of the NER-Informed model, when focusing only on their top-$k$ outputs.\footnote{Both models typically output between 1--3 entity links ranked according to their likelihood.} We aggregate these statistics at the sentence level over the whole corpus.
We focus on two datasets, namely WikiANN and MasakhaNER, summarizing the results in Figure~\ref{fig:ner_inf}.\footnote{An extensive results table is available in Appendix~\ref{app:ner_inf}.}

Comparing the general performance between these two datasets, it is clear that general agreement is decent. 
In 7 Out of 9 typologically diverse languages from WikiANN, more than 60\% top-1 entities are linked by both models. The African languages from MasakhaNER are low-resource ones yielding less than 40\% EL agreement to English in all cases. Given that most of these languages have not been included in the pre-training of BART (the model mGENRE is based on), we expect that using AfriBERTa~\cite{oguejismall} or similar models in future work would yield improvements.


\paragraph{Effect on downstream maps} 
We compare the dataset maps we obtain using NER-Relaxed and NER-Informed (using gold annotations) models in our pipeline for the MasakhaNER dataset. Overall, the maps are very similar. An example visualization of the two maps obtained for Swahili is in Figure~\ref{fig:comparison_maps} in Appendix~\ref{app:comparison}. 

The NER-Informed model produces slightly fewer entities overall (likely exhibiting higher precision for lower link recall) but there are minimal differences on the representativeness measures e.g., the in-country percentage changes from 15.3\% (NER-Informed) to 16.9\% (NER-Relaxed).
We can compare the distributions of the top-$k$ countries obtained with the two models using Ranked Biased Overlap \cite[RBO; higher is better;][]{webber2010similarity}.\footnote{See Appendix~\ref{app:comparison} on why this metric is appropriate.} 
The results for varying values for $k$ (top-$k$ countries) are presented in Table~\ref{tab:rank_rbo} in Appendix~\ref{app:comparison}. We overall obtain very high RBO values ($>.8$ for $k=10$) for all language portions and all values of $k$. For example for~8 of the~10 MasakhNER languages the two models almost completely agree on the top-10 countries with only slight variations in their ranking. Dholuo and Amharic are the ones exhibiting the worse overlap (but still $>.5$ RBO). 

\section{Conclusion}
We present a recipe for visualizing how representative NLP datasets are with respect to the underlying language speakers. 
We plan to further improve our tool\footnote{To be released as a Python package.} 
by making NER/EL models more robustly handle low-resource languages. We will also expand our dataset and task coverage, to get a broader overview of the current utility of NLP systems.

\section*{Acknowledgements}
This work is generously supported by NSF Awards 2040926 and 2125466.

\bibliography{custom,anthology}
\bibliographystyle{acl_natbib}

\clearpage
\newpage
\appendix

\section{Responsible NLP Notes}
\label{app:responsible}

We use this section to expand on potential limitations and risks of this work.

An inherent limitation of this work is that many datasets are constructed with the goal of answering scientific questions -- not necessarily to be used to build NLP systems that serve language users. If our tool is applied without the assumptions behind dataset construction in mind, it might lead to undue criticisms of existing datasets. It us also important to reiterate that no tool, including ours, will ever be 100\% accurate, so our tool should be used as \textit{an indicator} of the cultural representativeness of language datasets, not as a tool that can provide definitive answers.

All scientific artifacts used in this paper are publicly available under permissive licenses for fair use. We are not re-distributing any data or code, beyond the code that we wrote ourselves (which will be released under a CC-0 license) and the additional annotations on top of the existing datasets which map the datasets to Wikidata entries (Wikidata data are also available under a CC-0 license). Our use of our data is consistent with their intended use.

\section{Related Work}
\label{app:related}

Effective measurement of dataset quality is an aspect of fast-growing significance. Training large language models require huge amount of data and as a result, the inference generated by these pretrained language model as well as the fine-tuned models often show inherent data bias. In a recent work~\cite{swayamdipta-etal-2020-dataset}, the authors present how data-quality aware design-decision can improve the overall model performance. They formulated categorization of data-regions based on characteristics such as out-of-distribution feature, class-probability fluctuation and annotation-level discrepancy. 

Usually, multilingual datasets are collected from diverse places. So it is important to assess whether the utility of these datasets are representative enough to reflect upon the native speakers. We find the MasakhaNER~\cite{adelani2021masakhaner} is one such dataset that was collected from local sources and the data characteristics can be mapped to local users as a result. In addition, language models often requires to be truly language-agnostic depending on the tasks, but one recent work shows that, the current state-of-the-art language applications are far from achieving this goal~\cite{joshi-etal-2020-state}. The authors present quantitative assessment of available applications and language-resource trajectories which turns out not uniformly distributed over the usefulness of targeted users and speakers from all parts of the world. 

Linking dataset entities to geospatial concept is one integral part of our proposed methodology. Ongoing geospatial semantics research mostly focuses on extracting spatial and temporal entities~\cite{ijgi9030146, INR-034}. The usual approach is to first extract geo-location concepts (i.e. geotagging) from semi-structured as well as unstructured data and then linking those entities to location based knowledge ontology (i.e. geocoding). In ~\cite{article-prag}, the authors propose a task-metric-evaluation framework to evaluate existing NER based geoparsing methods. The primary findings suggest that NER based geo-tagger models in general rely on instant word-sense while avoiding contextual information.

One important aspect of our study is the evaluation of cross-lingual consistency while performing multilingual NER or El tasks. In~\cite{bianchi2021language}, the authors focus on the consistency evaluation of language-invariant properties. In an ideal scenario, the properties should not be changed via the language transformation models but commercially available models are not prone to avoid domain dependency.

\section{Dataset Statistics}
\label{app:d_stat}
\begin{table*}[]
    \centering
    \small
\begin{tabular}{p{0.1\textwidth}|p{0.08\textwidth}|p{0.4\textwidth}|p{0.1\textwidth}|p{0.1\textwidth}}
\toprule
    \textbf{ Dataset }& \textbf{Data-split} &                                                                                                                                                                                                                                                                                                                                                                                                                                                                                                                     \textbf{ Languages} &  \textbf{Language count} &  \textbf{Sentence count} \\

\midrule
        WikiANN &   train    &  russian, polish, kazakh, bulgarian, finnish, ukrainian, afrikaans, hindi, yoruba, hungarian, dutch-flemish, korean, persian, japanese, javanese, portuguese, hebrew, arabic, spanish-castilian, bengali, urdu, indonesian, tamil, english, malayalam, tagalog, basque, thai, german, romanian- moldavian-moldovan, chinese, telugu, azerbaijani, quechua, modern-greek, turkish, marathi, georgian, estonian, italian, panjabi, burmese, french, gujarati, malay, lithuanian, swahili, vietnamese &          48 &      658600 \\
        \midrule
      TyDi-QA &   train    &                                                                                                                                                                                                                                                                                                                                                                                                                    english, korean, japanese, telugu, russian, thai, arabic, finnish, bengali, swahili, indonesian &          11 &      166905 \\
      \midrule
   MasakhaNER &   train    &                                                                                                                                                                                                                                                                                                                                                                                                                               igbo, wolof, nigerian pidgin, kinyarwanda, amharic, hausa, yoruba, ganda, swahili, dholuo &          10 &       12906 \\
   \midrule
 SQuAD &   train    &                                                                                                                                                                                                                                                                                                                                                                                                                                                                                                                        english &           1 &      130319 \\
 \midrule
        MLQA &  dev, test     &                                                                                                                                                                                                                                                                                                                                                                                                                                                 english, simplified chinese, german, arabic, spanish, hindi, vietnamese &           7 &       12738 \\
        \midrule
\iffinal    WMT NEWS &   dev, test    &                                                                                                                                                                                                                                                                                                                              polish, kazakh, finnish, xhosa, hindi, japanese, bengali, tamil, zulu, romanian; moldavian; moldovan, chinese, estonian, french, gujarati, inuktitut, lithuanian, turkish, latvian, dholuo, english &          20 &      126972 \\
    \midrule
    \fi
    Natural Questions &   train    &                                                                                                                                                                                                                                                                                                                                                                                                                                                                                                                        english &           1 &      307373 \\
\bottomrule
\end{tabular}
    \caption{Statistics of the datasets we study.}
    \label{tab:data_stat}
\end{table*}
See details in Table~\ref{tab:data_stat}.

\section{Geographical Breakdown of Models Performance}
\label{app:geo_f1}
\begin{table*}[t]
\small
    \centering
\begin{tabular}{l|llllll}
\toprule
{} &  europe &  asia &  africa &  americas &  history &  oceania \\
\midrule
swahili    &    (80.3, 88.9) &  (64.1, 83.4) &    (75.5, 81.4) &      (88.1, 89.3) &     (83.3, 100) &     (86.5, 81.2) \\
bengali    &    (60.0, 79.6) &  (71.0, 79.5) &   - &       - &    (100, 100) &      (0, 100) \\
arabic     &    (65.2, 79.0) &  (74.5, 82.6) &    (72.3, 79.0) &      (82.4, 82.6) &     (36.3 65.6) &    (100, 100) \\
korean     &    (19.3, 23) &  (30.4, 36.5) &     (0, 0) &      (23.9, 24.6) &     (42.9, 52.4) &      - \\
english    &    (74.7, 89.2) &  (84.0, 80.2) &    (60.0, 60.0) &      (75.6, 82.9) &    (100, 100) &     (93.3, 93.3) \\
indonesian &    (79.4, 88.5) &  (75.3, 84) &    (80, 100) &      (79.9, 84.7) &     (83.3, 66.7) &     (33.3, 33.3) \\
russian    &    (65.1, 80.1) &  (59.6, 79.1) &    (64.9, 67.8) &      (67.8, 81.8) &     (47.2, 72.3) &     (76.8, 66.7) \\
telugu     &    (63.7, 77.3) &  (45.9, 77.9) &    (83.3, 83.3) &      (34.5, 65.7) &    (100, 100) &     (66.7, 100) \\
finnish    &    (73.4, 81) &  (86.2, 88.9) &    (81, 91.7) &      (75.9, 83) &     (67.7, 74.7) &      - \\
\bottomrule
\end{tabular}
    \caption{Detailed Breakdown of area-based performance (f1 score) of two trained QA models (TyDi-QA, SQuAD). Evaluation is performed on TyDi-QA development set (gold task).}
    \label{tab:geo_f1}
\end{table*}
See details in Table~\ref{tab:geo_f1}.

\section{NER-Informed vs NER-Relaxed Models}
\label{app:ner_inf}
In this section, we report the detailed results (see Table~\ref{tab:ner_inf}) from our experiment with using intermediate NER model vs skipping this step. 
\begin{table*}[t]
\small
    \centering
\begin{tabular}{l|lll|l}
\toprule
 \textbf{Language} &                 \textbf{ k=1} &                   \textbf{k=2 }&                  \textbf{ k=3} &     \textbf{Dataset} \\
\midrule
hin &    (4239, 761, 0.85) &    (6765, 2717, 0.71) &    (8377, 4436, 0.65) &     \multirow{ 9}{*}{WikiANN} \\
cmn &  (9354, 10646, 0.47) &   (16015, 23899, 0.4) &  (21835, 37346, 0.37) &      \\
jpn &  (6739, 13259, 0.34) &   (12148, 27820, 0.3) &  (17220, 42463, 0.29) &      \\
rus &  (15325, 4675, 0.77) &  (24663, 13989, 0.64) &  (31520, 23051, 0.58) &      \\
est &  (16687, 3313, 0.83) &   (24413, 10536, 0.7) &  (28146, 16459, 0.63) &      \\
ben &    (9575, 425, 0.96) &   (15759, 2541, 0.86) &    (20106, 4930, 0.8) &      \\
que &       (82, 18, 0.82) &       (124, 48, 0.72) &       (159, 72, 0.69) &      \\
tur &  (14206, 5794, 0.71) &  (21165, 14999, 0.59) &  (25053, 23597, 0.51) &      \\
jav &       (78, 22, 0.78) &       (103, 67, 0.61) &      (113, 101, 0.53) &      \\
\midrule
pcm &     (549, 994, 0.36) &     (955, 2033, 0.32) &    (1217, 3030, 0.29) &  \multirow{10}{*}{MasakhaNER} \\
kin &     (593, 952, 0.38) &     (924, 1988, 0.32) &    (1112, 2853, 0.28) &   \\
wol &     (242, 534, 0.31) &     (350, 1158, 0.23) &      (435, 1692, 0.2) &   \\
hau &    (417, 1178, 0.26) &     (747, 2333, 0.24) &     (941, 3402, 0.22) &   \\
ibo &    (494, 1093, 0.31) &     (834, 2225, 0.27) &    (1056, 3257, 0.24) &   \\
amh &     (117, 1088, 0.1) &     (210, 2184, 0.09) &     (289, 3198, 0.08) &   \\
swa &     (499, 1175, 0.3) &     (819, 2445, 0.25) &    (1007, 3678, 0.21) &   \\
lug &     (283, 824, 0.26) &     (486, 1657, 0.23) &     (644, 2362, 0.21) &   \\
yor &     (430, 894, 0.32) &     (673, 1909, 0.26) &     (839, 2893, 0.22) &   \\
luo &     (122, 428, 0.22) &       (207, 844, 0.2) &     (264, 1184, 0.18) &   \\
\bottomrule
\end{tabular}
    \caption{Breakdown of entity extraction count while using NER-informed model. Here for each top k extracted entities, the triplet is the aggregated value of (count of common entities extracted by both ner-informed and ner-relaxed models, count of entities only extracted by ner-relaxed models, ratio of common entity count and total top-k extract by ner-relaxed model )}
    \label{tab:ner_inf}
\end{table*}

\subsection{Comparison of NER-Informed and NER-Relaxed Maps}
\label{app:comparison}
\begin{figure*}[t]
    \centering
    \begin{tabular}{cc}
        \includegraphics[width=.48\textwidth]{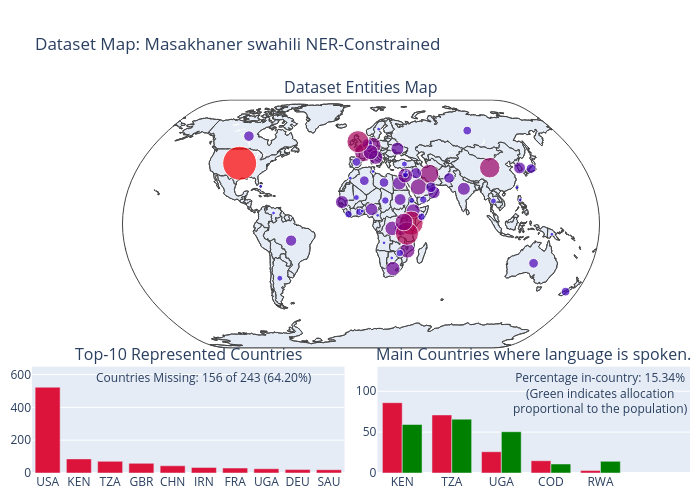} &  \includegraphics[width=.48\textwidth]{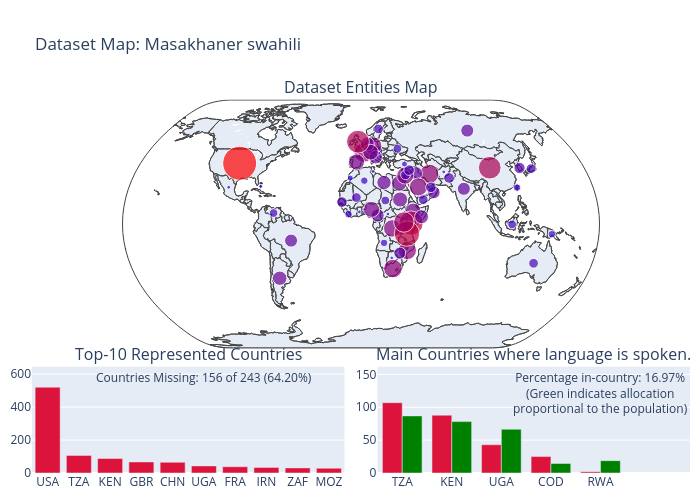} \\
        (a) Swahili NER-Informed & Swahili NER-Relaxed \\
    \end{tabular}
    
    \caption{The dataset maps obtained by NER-Informed and NER-Relaxed are very similar, with very small differences in the representativeness measures.}
    \label{fig:comparison_maps}
\end{figure*}

This experiment was performed on MasakhaNER data. See Figure~\ref{fig:comparison_maps} for example maps in Swahili. 
The distributions of the top-$k$ countries we obtain with the two models (one using the gold NER annotations for NEL and one using our NER-relaxed approach) are compared using Ranked Biased Overlap (RBO; higher is better) \cite{webber2010similarity}, a metric appropriate for computing the weighted similarity of disjoint rankings. We choose a ``weighted" metric because we care more about having similar results in the top-$k$ countries (the ones most represented) so that the metric is not dominated by the long tail of countries that may have minimal representation and thus similar rank. We also need a metric that can handle disjoint rankings, since there's no guarantee that the top-$k$ countries produced by the processes using different models will be different.\footnote{Metrics like Kendall's $\tau$ would suffer from both issues.}

The results for varying values for $k$ (top-$k$ countries) are presented in Table~\ref{tab:rank_rbo}. We overall obtain very high RBO values ($>.75$) for all language portions and all settings. 

\begin{table*}[t]
    \centering
    \begin{tabular}{l|ccccccccc}
        \toprule
        Dataset & \multicolumn{9}{c}{Rank Biased Overlap (RBO) for top-$k$ ranked countries with $k$=} \\
        Portion & 1 & 2 & 3 & 5 & 10 & 20 & 50 & 100 & 200  \\
        \midrule
Amharic & 0.00 & 0.25 & 0.50 & 0.57 & 0.53 & 0.51 & 0.59 & 0.65 & 0.76 \\
Yoruba & 1.00 & 1.00 & 1.00 & 0.87 & 0.82 & 0.87 & 0.85 & 0.83 & 0.87 \\
Hausa & 1.00 & 1.00 & 1.00 & 0.87 & 0.80 & 0.80 & 0.83 & 0.83 & 0.88 \\
Igbo & 1.00 & 1.00 & 1.00 & 0.96 & 0.89 & 0.82 & 0.79 & 0.79 & 0.86 \\
Kinyarwanda & 1.00 & 0.75 & 0.83 & 0.86 & 0.91 & 0.89 & 0.83 & 0.80 & 0.86 \\
Luganda & 1.00 & 0.75 & 0.83 & 0.81 & 0.81 & 0.82 & 0.78 & 0.76 & 0.83 \\
Dholuo & 1.00 & 0.75 & 0.83 & 0.77 & 0.66 & 0.58 & 0.57 & 0.62 & 0.76 \\
Nigerian Pidgin & 1.00 & 1.00 & 1.00 & 0.95 & 0.91 & 0.90 & 0.89 & 0.86 & 0.90 \\
Wolof & 1.00 & 1.00 & 1.00 & 0.96 & 0.85 & 0.77 & 0.68 & 0.70 & 0.81 \\
Swahili & 1.00 & 0.75 & 0.83 & 0.90 & 0.89 & 0.89 & 0.84 & 0.85 & 0.90 \\
\midrule
Average & 0.90 & 0.82 & 0.88 & 0.85 & 0.81 & 0.79 & 0.76 & 0.77 & 0.84 \\
\bottomrule
    \end{tabular}
    \caption{Rank Biased Overlap (RBO; higher is better) for the top-$k$ ranked countries obtained by a NER-Informed and a NER-Relaxed model on the MasakhaNER datasets.}
    \label{tab:rank_rbo}
\end{table*}


\section{On the Cross-Lingual Consistency of NER/EL Models}
\label{sec:consistency}

\paragraph{Definition}
\citet{bianchi2021language} in concurrent work point out the need to focus on consistency evaluation of \textbf{language-invariant properties (LIP)}: properties which should not be changed via language transformation models. They suggest LIPs include meaning, topic, sentiment, speaker demographics, and logical entailment We propose a definition tailored to entity-related tasks: cross-lingual consistency is the desirable property that two parallel sentences in two languages, which should in principle use the same named entities (since they are translations of each other), are actually tagged with the same named entities. 

\subsection{NER Experiments}
\paragraph{Models} We study two 
models: SpaCy~\cite{spacy2}: a state-of-art monolingual library that supports several core NLP tasks; and a mBERT-based NER model trained on datasets from WikiANN using the transformers library~\cite{wolf2020huggingfaces}.

\paragraph{Training}
To task-tune the mBERT-based model on the NER task we use the WikiANN dataset with data from the four languages we study: Greek (el), Italian (it), 
Chinese (zh), and English (en). 

\begin{table}[t]
    \centering
    \begin{tabular}{lccc}
    \toprule
        Model & Greek & Italian & Chinese  \\
    \midrule
        \small Monolingual (SpaCy) & 8.6 & 3.1 & 14.1 \\
    \midrule
        mBERT & \textbf{53.4} & \textbf{62.9} & \textbf{25.5} \\
    \bottomrule
    \end{tabular}
    \caption{Using a multilingual NER model leads to significantly higher consistency tested on Eng--X data.}
    \label{tab:ner_results}
    \vspace{-1em}
\end{table}

\paragraph{Evaluation}
To evaluate cross-lingual consistency, ideally one would use parallel data where both sides are annotated with named entities. What we use instead, since such datasets do not exist to the best of our knowledge, is `silver' annotations over parallel data. We start with unannotated parallel data from the WikiMatrix dataset~\cite{schwenk-etal-2021-wikimatrix} and we perform NER on both the English and the other language side, using the respective language model for each side.  

In the process of running our experiments, we identified some sources of noise in the WikiMatrix dataset (e.g. mismatched sentences that are clearly not translations of each other). Thus, we calculated the average length ratio between two matched sentences, and discarded data that diverged by more than one standard deviation from the mean ratio, in order to keep $95$\% of the original data that are more likely to indeed be translations of each other.

We use the state-of-the-art AWESOME-align tool~\cite{dou2021word} 
\iffinal as well fast-align~\cite{dyer-etal-2013-simple} 
\fi
to create word-level links between the words of each English sentence to their corresponding translations. 
Using these alignment links for cross-lingual projection~\cite[\textit{inter alia}]{pado2009cross,tiedemann2014rediscovering,ni2017weakly} allows us to calculate cross-lingual consistency, measuring the portion of labels that agree following projection. In particular, we use the cross-lingual projections from the English side as `correct' and measure precision, recall, and F-score against them.

\paragraph{Results}
\iffinal
In preliminary experiments we found that, consistently with the literature, AWESOME-align performed generally better than fast-align, hence for the remainder of our experiments we only use AWESOME-align.  
\fi

For the three languages we study, the cross-lingual consistency of the monolingual SpaCy models is really low, with scores of $8.6$\% for Greek--English, $3.1$\% for Italian--English and $14.1$\% for Chinese--English. 
The SpaCy models are independently trained for each language and can produce 18 fine-grained NE labels e.g. distinguishing dates from time, or locations to geopolitical entities. As such, there was no a priori expectation for high cross-lingual consistency. Nevertheless, these extremely low scores reveal deeper differences, such as potentially widely different annotation protocols across languages.\footnote{We note that our evaluation does focus only on labels shared between models/languages.}

For the mBERT-based model we again label both sides of the parallel data, but now evaluate only on locations (LOC), organizations (ORG) and persons (PER) (the label types present in WikiANN).
The mBERT models have significantly higher cross-lingual consistency: on the same dataset as above, we obtain $53.4$\% for Greek to English, $62.9$\% for Italian to English and $25.5$\% for Chinese to English. 

\paragraph{Discussion} To further understand the source of cross-lingual discrepancies, we performed manual analysis of 400 Greek-English parallel sentences where the mBERT-based model's outputs on Greek and the projected labels through English disagreed.\footnote{We chose this language pair because one of the authors is a fluent speaker of both languages.} 
We sampled 100 sentences where the English-projected label was \texttt{0} but the Greek one was 
\texttt{LOC} (location), 100 sentences with English-projected as \texttt{LOC} but Greek as \texttt{0}, and similarly for persons (\texttt{PER}). 

We performed annotation using the following schema:
\begin{itemize}[leftmargin=*]
    \item Greek wrong: for cases where only the English-side projected labels are correct
    \item English wrong: for cases where the English-side projected labels are wrong but the Greek-side are correct
    \item both wrong: for cases where the labels on both sides are incorrect
    \item alignment wrong: for cases where the two aligned phrases are not translations of each other, so we should not take the projected labels into account nor compare against them.
    \item all correct: both sides as well as the alignments are correctly tagged (false negatives).
\end{itemize}

Encouragingly, the entity alignments were wrong in less than 10\% of the parallel sentences we manually labelled. This means that our results are quite robust: a 10\%-level of noise cannot account for an almost 50\% lack of consistency on the Greek-English dataset.\footnote{It does provide a potential upper bound of around 90\% on the consistency we should expect to find.} Hence, the system definitely has room for improvement. A second encouraging sign is that less than 2\% of the cases were in fact false negatives, i.e. 
due to the phrasing of the translation 
only one of the two sides actually contained an entity. 

Going further, we find that mistakes vary significantly by label type. In about 75\% of the \texttt{0-LOC} cases it was the Greek-side labels that were wrong in outputting \texttt{LOC} tags. A common pattern (about 35\% of these cases) was the Greek model tagging months as locations.  In the case of \texttt{0-PER} cases, 62\% of the errors were on the English side. A common pattern was the English-side model not tagging persons when they are the very first token in a sentence, i.e. the first token in {\small{\texttt{`Olga and her husband [...].'}}} Appendix~\ref{app:ner_discussion} extends this discussion with additional details and examples.

The above observations provide insights into NER models' mistakes, which we were able to easily identify by contrasting the models' predictions over parallel sentences. We argue this proves the utility and importance of also evaluating NER models against parallel data even without gold NER annotations.
Improving the NER cross-lingual consistency should in principle also lead to better NER models in general. Potential solutions could use a post-pretraining alignment-based fine-tuned mBERT model as the encoder for our data, or operationalize our measure of cross-lingual consistency into an objective function to optimize.\footnote{We leave this for future work, as it detracts off the main goal of this work (mapping datasets to the language users and measuring their representativeness).}

\subsection{Entity Linking Experiments}
We now turn to entity linking (EL), evaluating mGENRE's cross-lingual consistency  (under the NER-Relaxed setting, so the results below should be interpreted under this lens, as the NER-Informed --which we cannot run due to the lack of NER models for some languages-- could very well yield different results and analysis).

\paragraph{Dataset} We use parallel corpora from the WMT news translation shared tasks for the years 2014 to 2020 \cite{bojar-etal-2014-findings,bojar-etal-2015-findings,bojar-etal-2016-findings,bojar-etal-2017-findings,bojar-etal-2018-findings,barrault-etal-2019-findings,barrault-etal-2020-findings}. We work with 14 English-to-target language pairs, with parallel sentence counts in the range of around 1-5k.

\begin{figure}[t]
\centering
\includegraphics[width=7cm]{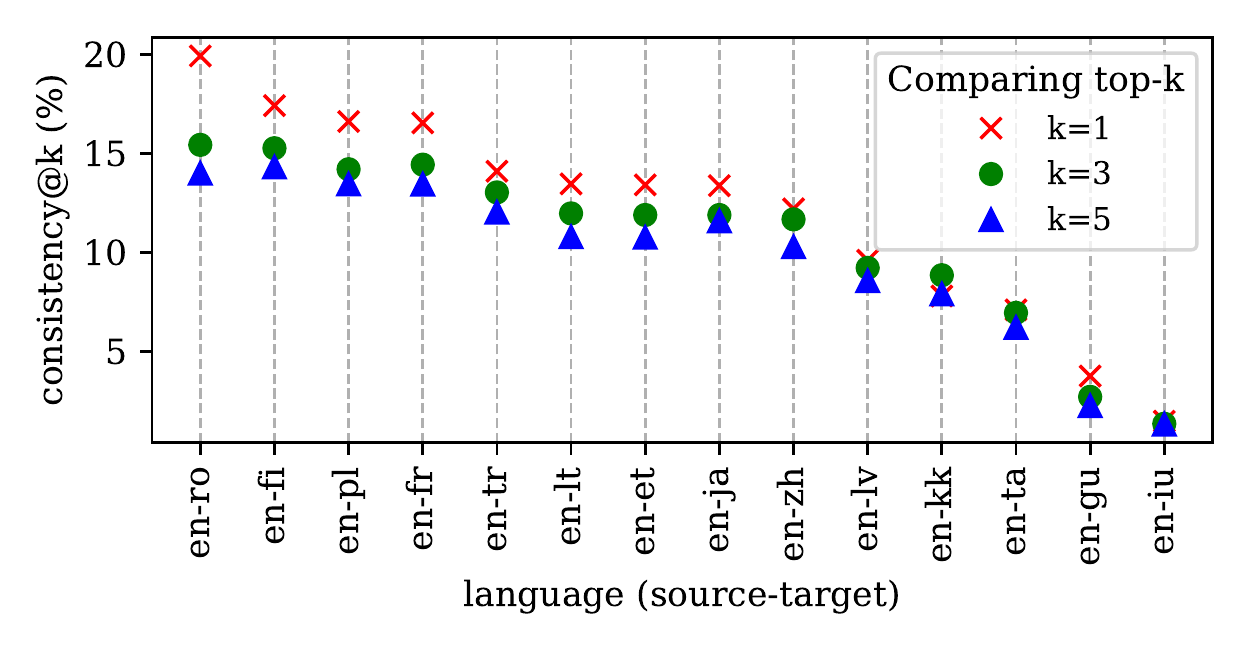}
\vspace{-1em}
\caption{The entity linking cross-lingual consistency is generally low across languages, but especially for low-resource language pairs like English to Inuktitut (iu), Gujarati (gu), or Tamil (ta).}
\label{fig:cross_cons}
\vspace{-1em}
\end{figure}

\paragraph{Evaluation} Unlike our NER experiment settings, we do not need word-level alignments to calculate cross-lingual consistency. We can instead compare the sets of the linked entities for both source and target sentences. 
\iffinal 
As before, we use mGENRE in a NER-Relaxed manner. 
In an ideal scenario, the output of the model over both source and target language sentences will include the same entity links, yielding a perfect cross-lingual consistency score of 1. 
\fi
In this manner, we calculate and aggregate sentence-level scores for the top-$k$ linked entities for $k=1,3,5$. In Figure~\ref{fig:cross_cons}, we present this score as a percentage, dividing the size of the intersection (of the source and target sentence outputs) by the number of source sentence entities. 

Additionally, in Table \ref{tab:cross_el}, we report the detailed cross-lingual consistency score percentages for 14 english-language source-target pairs from WMT news translation shared tasks~\cite{bawden-etal-2020-findings}.

\begin{table}[]

    \centering
\begin{tabular}{@{}clllc@{}}
\toprule
\textbf{src-tgt} & \begin{tabular}[c]{@{}l@{}}\textbf{k=1}\\  \textbf{\%}\end{tabular} & \begin{tabular}[c]{@{}l@{}}\textbf{k=3}\\  \textbf{\%}\end{tabular} & \begin{tabular}[c]{@{}l@{}}\textbf{k=5}\\  \textbf{\%}\end{tabular} & \begin{tabular}[c]{@{}l@{}}\textbf{sentence}\\       \textbf{count}\end{tabular} \\
\midrule
 en-ro &  19.91 &  15.42 &  13.98 &  1999 \\
 en-fi &  17.40 &  15.25 &  14.29 &  1500 \\
 en-pl &  16.60 &  14.19 &  13.43 &  2000 \\
 en-fr &  16.53 &  14.42 &  13.42 &  1500 \\
 en-tr &  14.09 &  13.02 &  12.01 &  1001 \\
 en-lt &  13.45 &  11.96 &  10.77 &  2000 \\
 en-et &  13.40 &  11.88 &  10.74 &  2000 \\
 en-ja &  13.36 &  11.88 &  11.57 &  1998 \\
 en-zh &  12.19 &  11.66 &  10.26 &  2002 \\
 en-lv &   9.59 &   9.21 &   8.55 &  2003 \\
 en-kk &   7.79 &   8.84 &   7.88 &  2066 \\
 en-ta &   7.09 &   6.94 &   6.19 &  1989 \\
 en-gu &   3.75 &   2.70 &   2.24 &  1998 \\
 en-iu &   1.47 &   1.34 &   1.31 &  5173 \\
\bottomrule
\end{tabular}
    \caption{Cross-lingual consistency score (\%) for top-k extracted and linked entities over all source language sentences.}
    \label{tab:cross_el}
\end{table}

\begin{table}[]

    \centering
\begin{tabular}{lrr}
\toprule
      \textbf{Entity category} &  \textbf{Common} &  \textbf{Source-only} \\
\midrule
         Unknown &    1720 &        16709 \\
      PERSON &    1358 &         5713 \\
         ORG &    1047 &         6911 \\
         GPE &     666 &         7379 \\
        NORP &     176 &         1895 \\
        DATE &     102 &         1427 \\
    CARDINAL &      78 &          565 \\
       EVENT &      77 &          777 \\
         LOC &      62 &          453 \\
 WORK\_OF\_ART &      20 &          133 \\
     PRODUCT &      15 &           91 \\
         FAC &      14 &          161 \\
    QUANTITY &       8 &           85 \\
        TIME &       6 &           43 \\
       MONEY &       4 &           14 \\
         LAW &       3 &          113 \\
    LANGUAGE &       3 &           80 \\
     ORDINAL &       2 &           90 \\
     PERCENT &       1 &            3 \\
     \midrule
    TOTAL &       5362 &            42642 \\
\bottomrule
\end{tabular}
    \caption{SpaCy NER ~\cite{spacy2} defined types and counts for consistent linked entities.}
    \label{tab:cross_el_etype}
\end{table}

\paragraph{Results}
As Figure~\ref{fig:cross_cons} shows, we obtain low consistency scores across all 14 language pairs, ranging from 19.91\% for English-Romanian to as low as 1.47\% for English-Inukitut ($k=1$). 
The particularly low scores for languages like Inuktitut, Gujarati, and Tamil may reflect the general low quality of mGENRE for such languages, especially because they use non-Latin scripts, an issue already noted in the literature~\cite{muller-etal-2021-unseen}.

The low percentage consistency scores for all languages makes it clear that mGENRE does not produce similar entity links for entities appearing in different languages. In future work, we plan to address this limitation, potentially by weighting linked-entities according to the cross-lingual consistency score when performing 
entity disambiguation in a multilingual setting.

\begin{figure}[t]
\centering
\includegraphics[width=7cm]{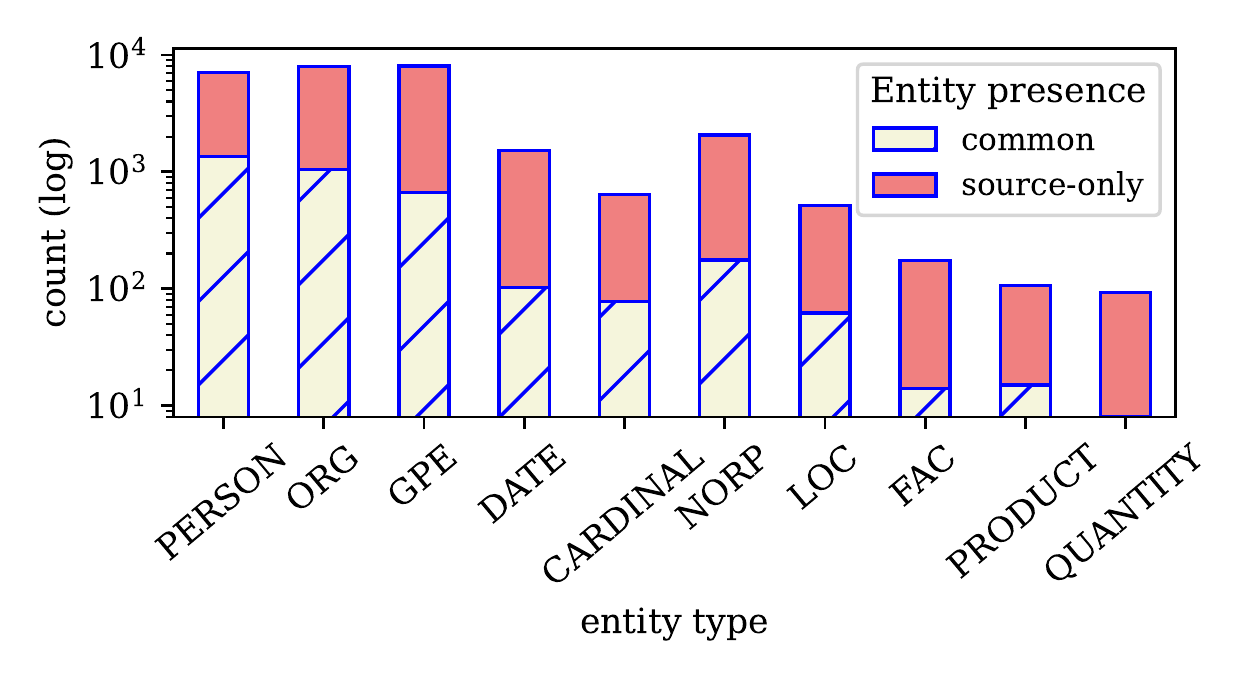}
\vspace{-1em}
\caption{Counts of linked entity types across all WMT language pairs. Notice the $y$-axis log-scale: 
many entities are linked differently on non-English input.}
\label{fig:cross_ent}
\vspace{-1em}
\end{figure}

\paragraph{Discussion}

We further analyze whether specific types of entities are consistently recognized and linked across language. We use SpaCy's English NER model to categorize all entities. 
Figure~\ref{fig:cross_ent} presents a visualization comparing consistent entity category counts to source-only ones. 

From Figure~\ref{fig:cross_ent}, it is clear that geopolitical entities (\texttt{GPE}) are the ones suffering the most from low cross-lingual consistency, with an order of magnitude less entities linked on both the English and the other language side. On the other hand, person names (\texttt{PER}) seem to be easier to link.
While the most common types of entities are \texttt{PERSON}, \texttt{ORG} (i.e. organization) and \texttt{GPE} (i.e. geopolitical entity), we found that the NER model still failed to correctly categorize entities like  (Surat, \texttt{Q4629}, \texttt{LOC}), (Aurangzeb, \texttt{Q485547}, \texttt{PER}). However, these entities were correctly linked by the NER-Relaxed pipeline, indicating its usefulness.
We hypothesize, and plan to test in future work, that a NER-Relaxed entity further regularized towards cross-lingual consistency will perform better than a NER-Informed pipeline, unless the NER component also shows improved cross-lingual consistency.

From Figure~\ref{fig:cross_ent}, it is clear that geopolitical entities (\texttt{GPE}) are the ones suffering the most from low cross-lingual consistency, with an order of magnitude less entities linked on both the English and the other language side. On the other hand, person names (\texttt{PER}) seem to be easier to link.
While the most common types of entities are \texttt{PERSON}, \texttt{ORG} (i.e. organization) and \texttt{GPE} (i.e. geopolitical entity), we found that the NER model still failed to correctly categorize entities like  (Surat, \texttt{Q4629}, \texttt{LOC}), (Aurangzeb, \texttt{Q485547}, \texttt{PER}). However, these entities were correctly linked by the NER-Relaxed pipeline, indicating its usefulness.
We hypothesize, and plan to test in future work, that a NER-Relaxed entity further regularized towards cross-lingual consistency will perform better than a NER-Informed pipeline, unless the NER component also shows improved cross-lingual consistency.

\section{Additional Dataset Maps}
\label{app:maps}

\begin{figure*}[t]
    \centering
    \begin{tabular}{cc}
    \multicolumn{2}{c}{\textbf{MasakhaNER Geographic Coverage}}\\
        \includegraphics[width=.45\textwidth]{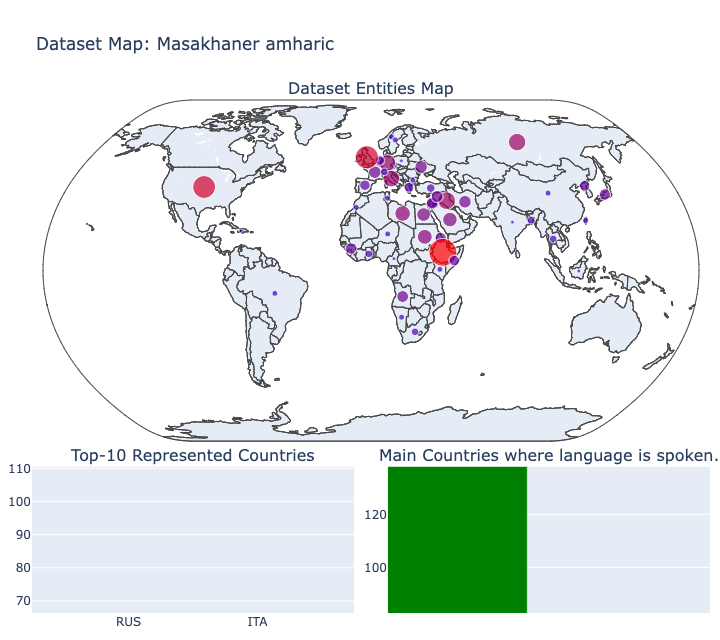} & 
        \includegraphics[width=0.45\textwidth]{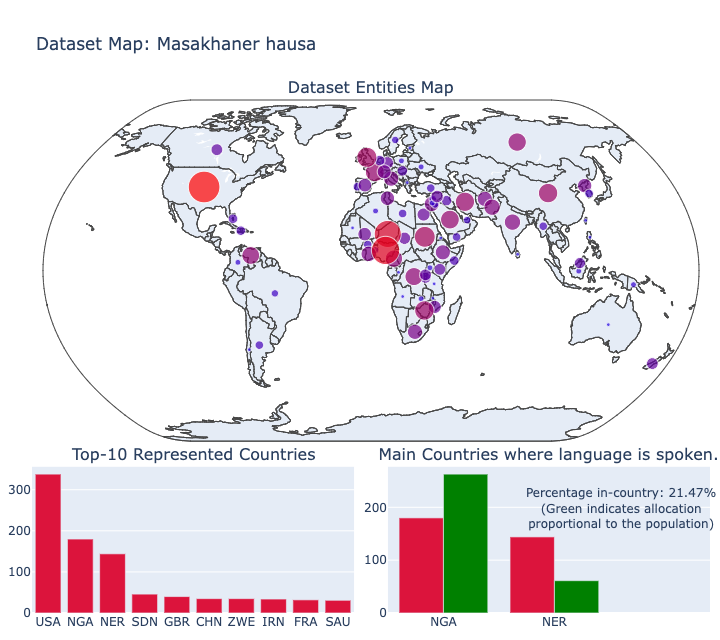}\\
        (a) Amharic & (b) Hausa \\
        \includegraphics[width=.45\textwidth]{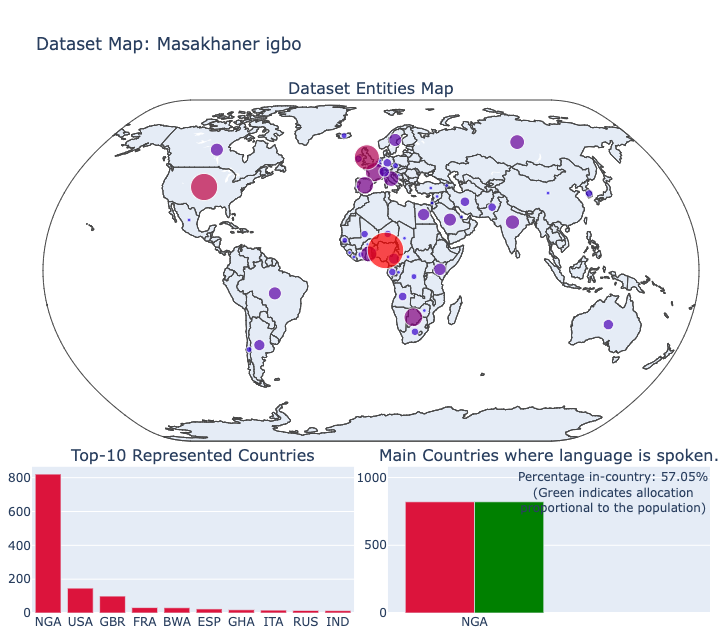} & 
        \includegraphics[width=0.45\textwidth]{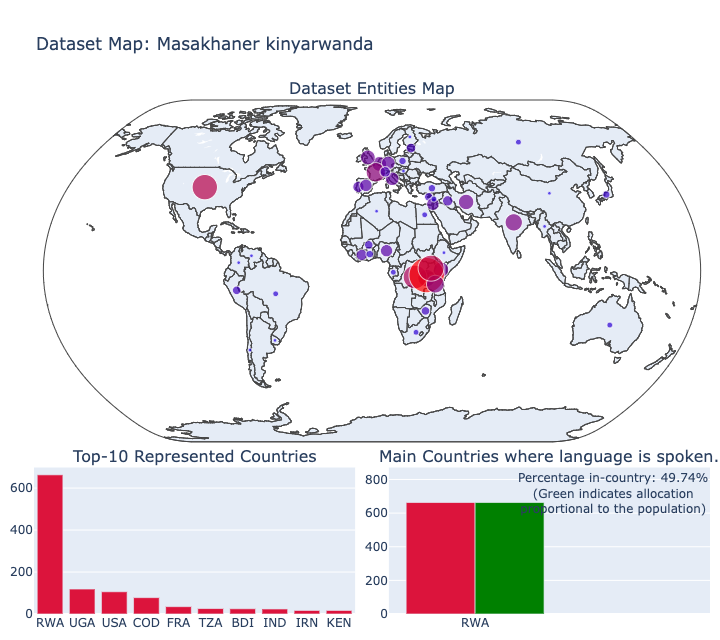}\\
        (c) Igbo & (d) Kinyarwanda \\
        \includegraphics[width=.45\textwidth]{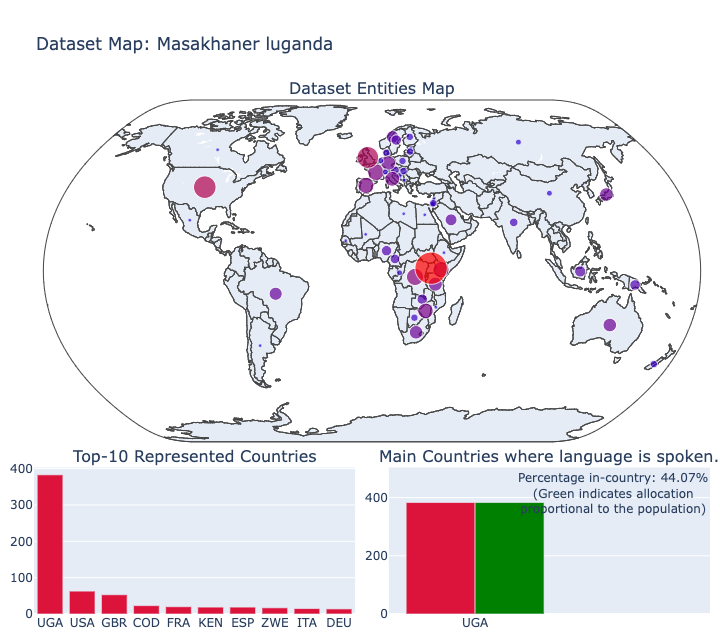} & 
        \includegraphics[width=0.45\textwidth]{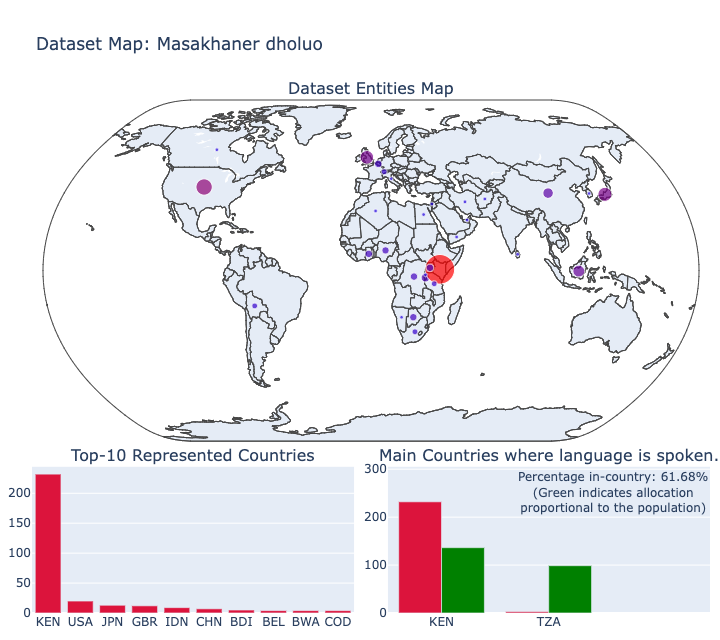}\\
        (e) Luganda & (f) Dholuo \\
    \end{tabular}
    \caption{MasakhaNER Geographic Distributions (Part 1).}
    \label{fig:masakhaner1}
\end{figure*}

\begin{figure*}[t]
    \centering
    \begin{tabular}{cc}
    \multicolumn{2}{c}{\textbf{MasakhaNER Geographic Coverage}}\\
        \includegraphics[width=.45\textwidth]{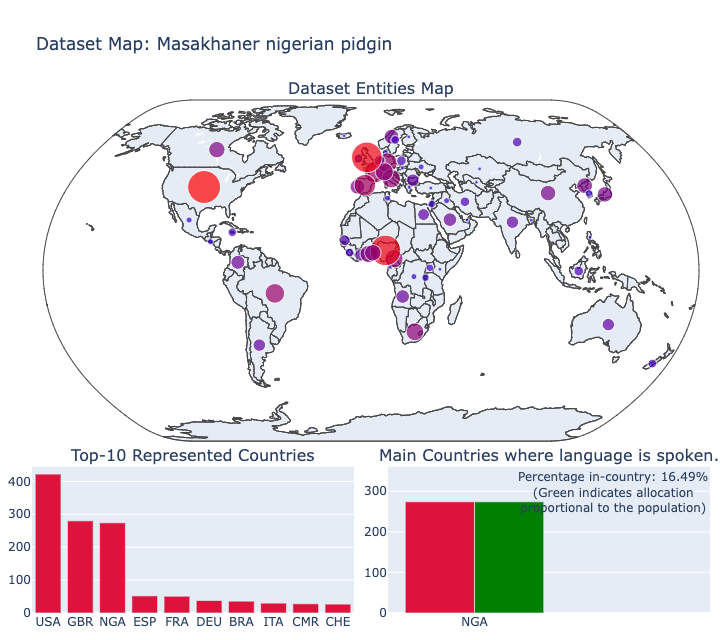} & 
        \includegraphics[width=0.45\textwidth]{plots/masakhaner-swa.png}\\
        (g) Nigerian English & (h) kiSwahili \\
        \includegraphics[width=.45\textwidth]{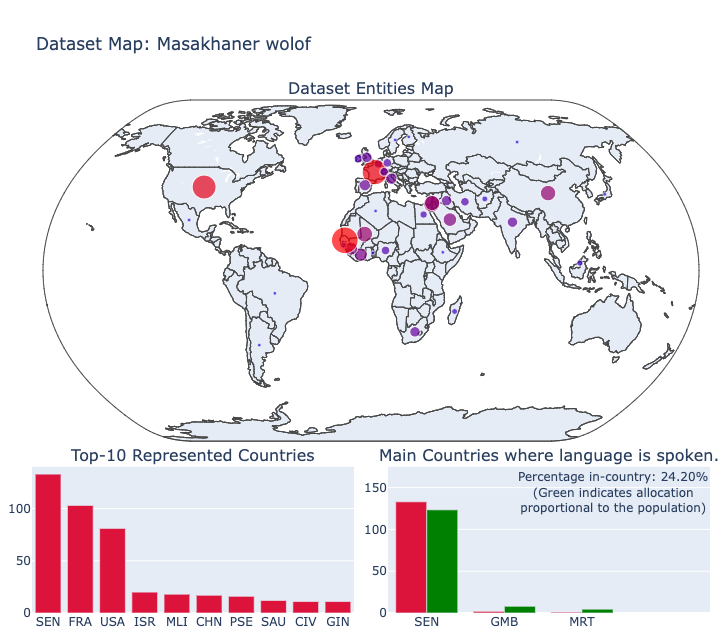} & 
        \includegraphics[width=0.45\textwidth]{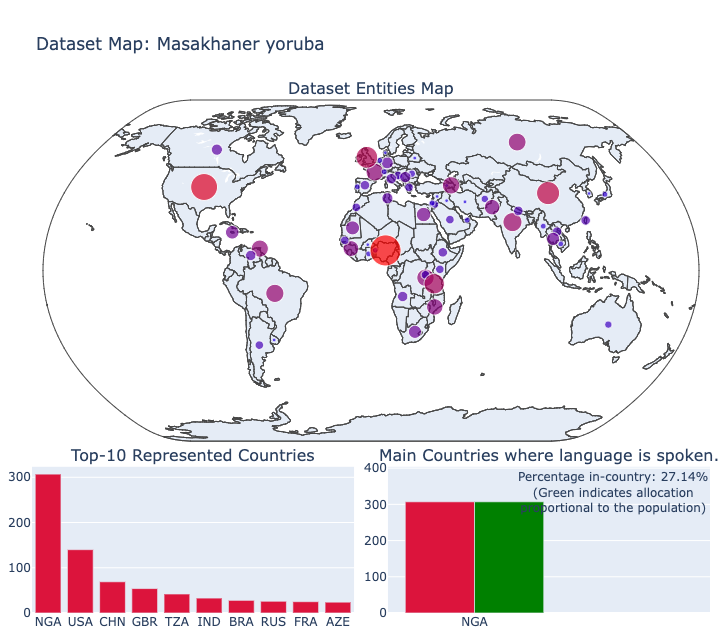}\\
        (i) Wolof & (j) Yoruba \\
    \end{tabular}
    \caption{MasakhaNER Geographic Distributions (Part 2).}
    \label{fig:masakhaner2}
\end{figure*}

\begin{figure*}[t]
    \centering
    \begin{tabular}{cc}
    \multicolumn{2}{c}{\textbf{TyDi-QA Geographic Coverage}}\\
        \includegraphics[width=.45\textwidth]{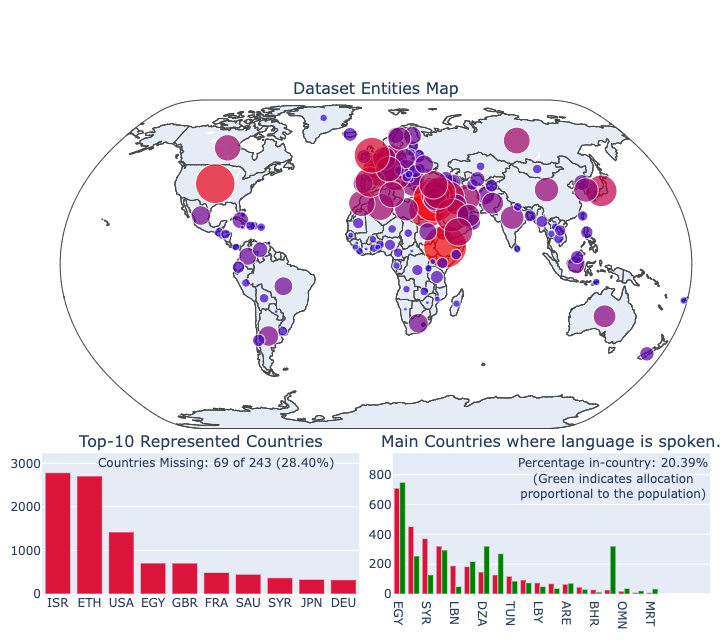} & 
        \includegraphics[width=0.45\textwidth]{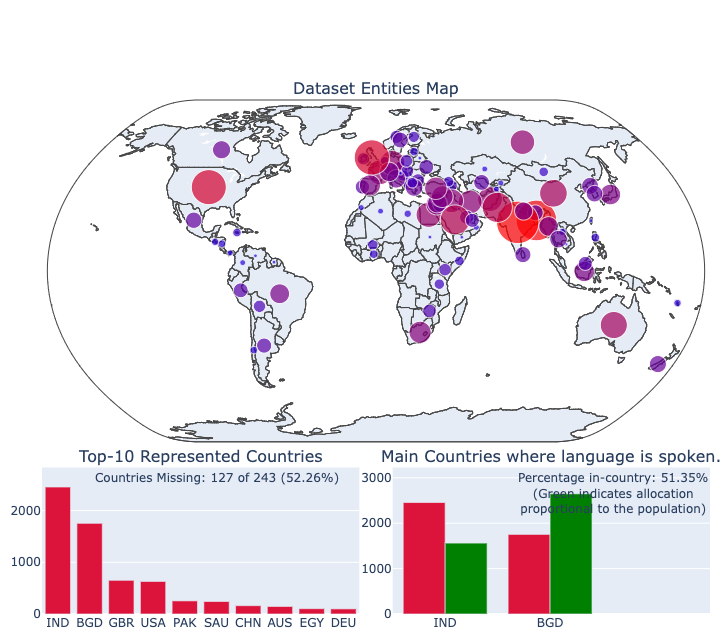}\\
        (a) Arabic & (b) Bengali \\
        \includegraphics[width=.45\textwidth]{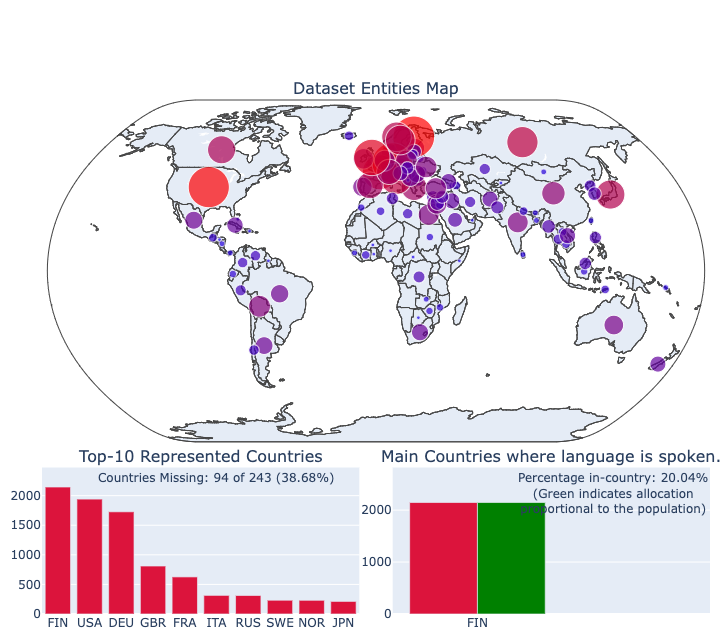} & 
        \includegraphics[width=0.45\textwidth]{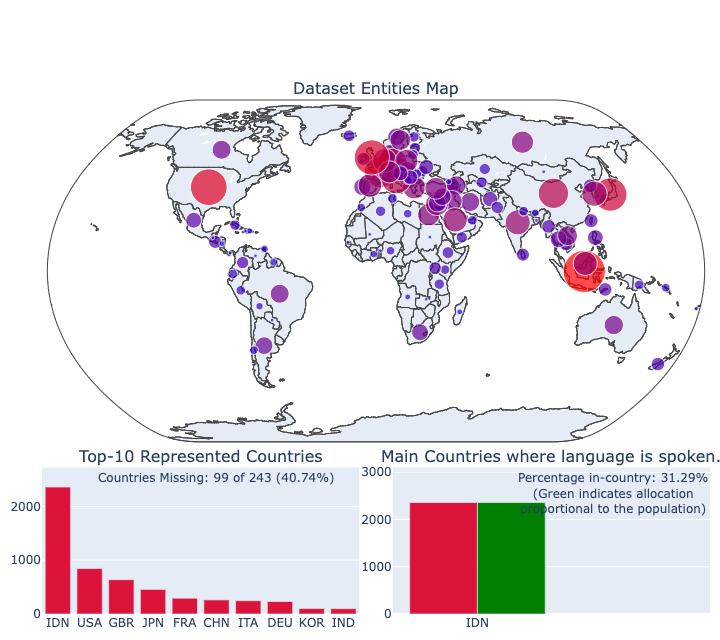}\\
        (c) Finnish & (d) Indonesian \\
        \includegraphics[width=.45\textwidth]{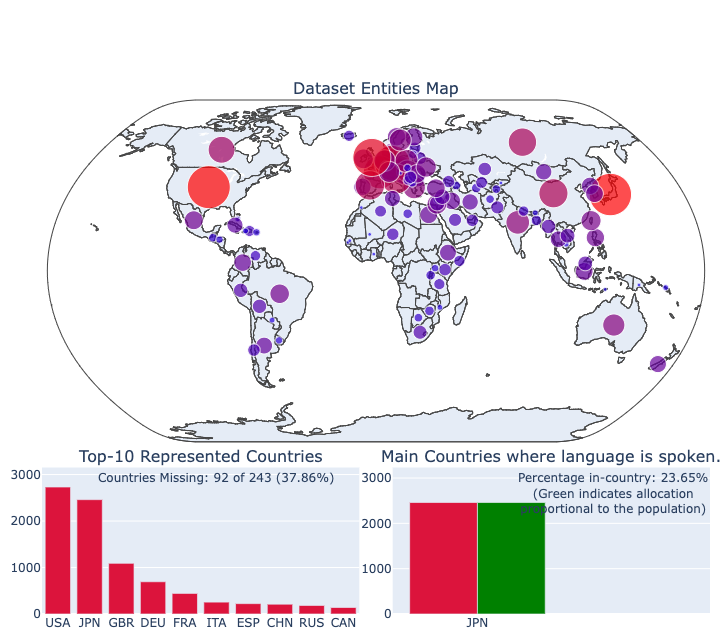} & 
        \includegraphics[width=0.45\textwidth]{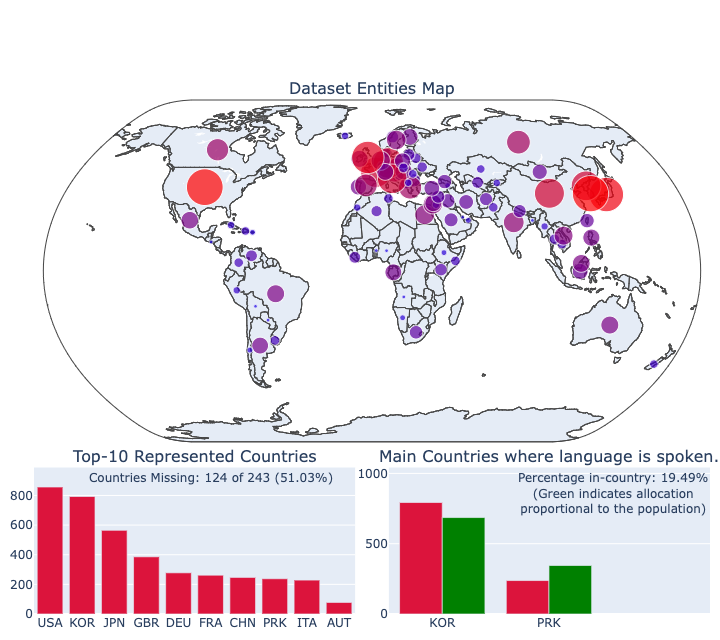}\\
        (e) Japanese & (f) Korean \\
    \end{tabular}
    \caption{TyDi-QA Geographic Distributions (Part 1).}
    \label{fig:tydiqa1}
\end{figure*}

\begin{figure*}[t]
    \centering
    \begin{tabular}{cc}
    \multicolumn{2}{c}{\textbf{TyDi-QA Geographic Coverage}}\\
        \includegraphics[width=.45\textwidth]{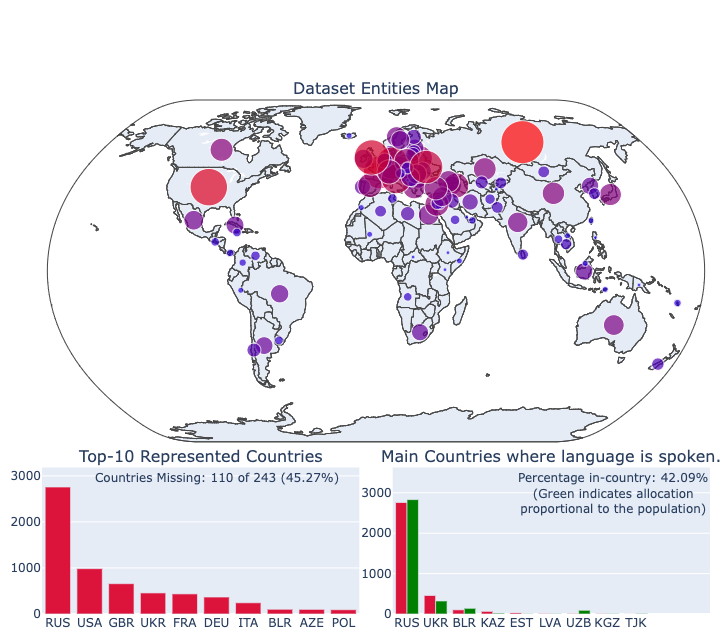} & 
        \includegraphics[width=0.45\textwidth]{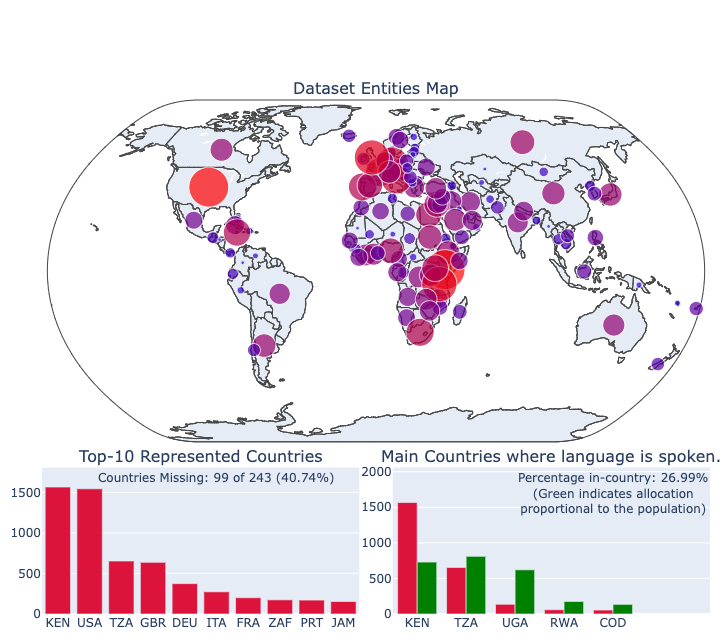}\\
        (g) Russian & (h) Swahili \\
        \includegraphics[width=.45\textwidth]{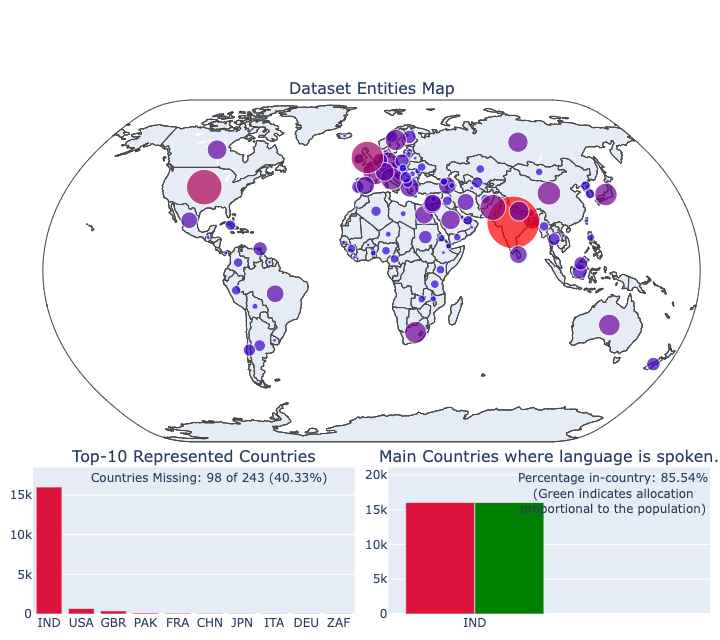} & 
        \includegraphics[width=0.45\textwidth]{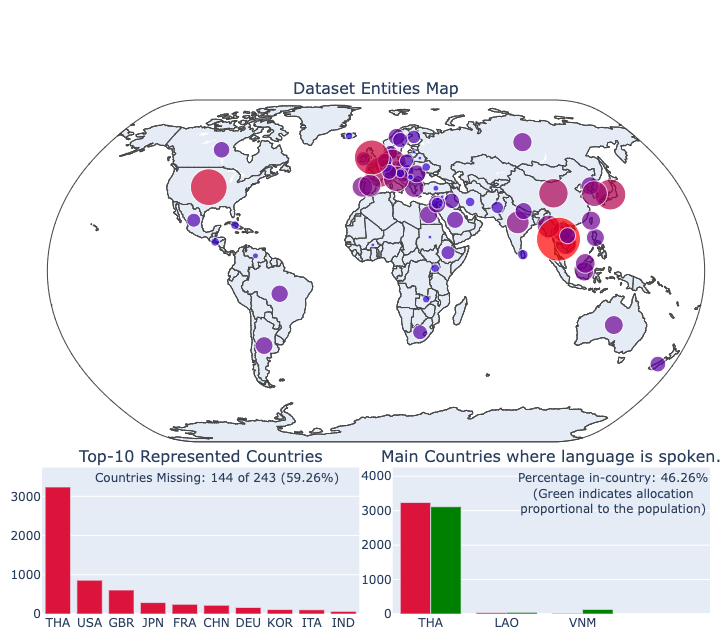}\\
        (i) Telugu & (j) Thai \\
    \end{tabular}
    \caption{TyDi-QA Geographic Distributions (Part 2).}
    \label{fig:tydiqa2}
\end{figure*}

\begin{figure*}[t]
    \centering
    \begin{tabular}{cc}
    \multicolumn{2}{c}{\textbf{Pan-X (WikiANN) Geographic Coverage}}\\
        \includegraphics[width=.45\textwidth]{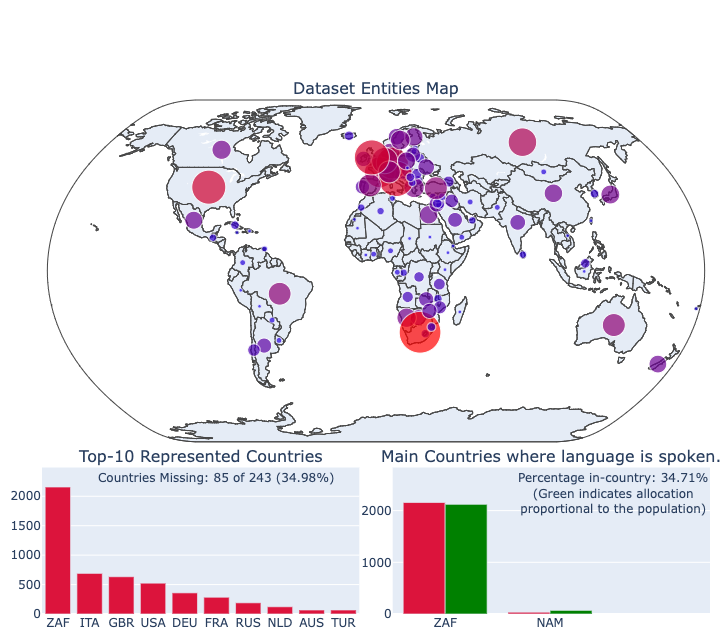} & 
        \includegraphics[width=0.45\textwidth]{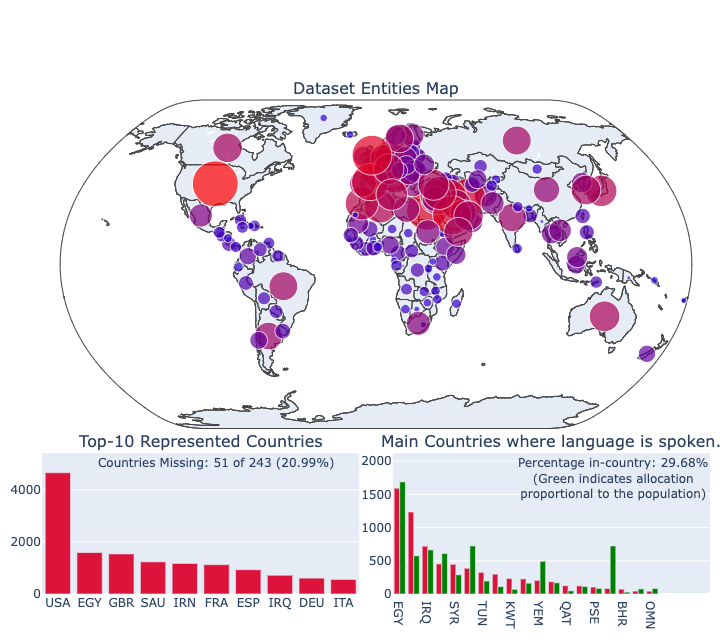}\\
        Afrikaans & Arabic \\
        \includegraphics[width=.45\textwidth]{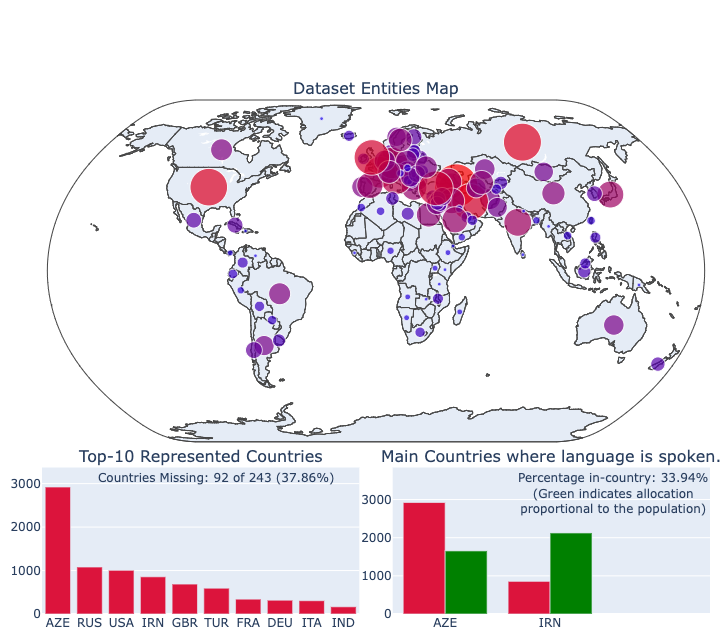} & 
        \includegraphics[width=0.45\textwidth]{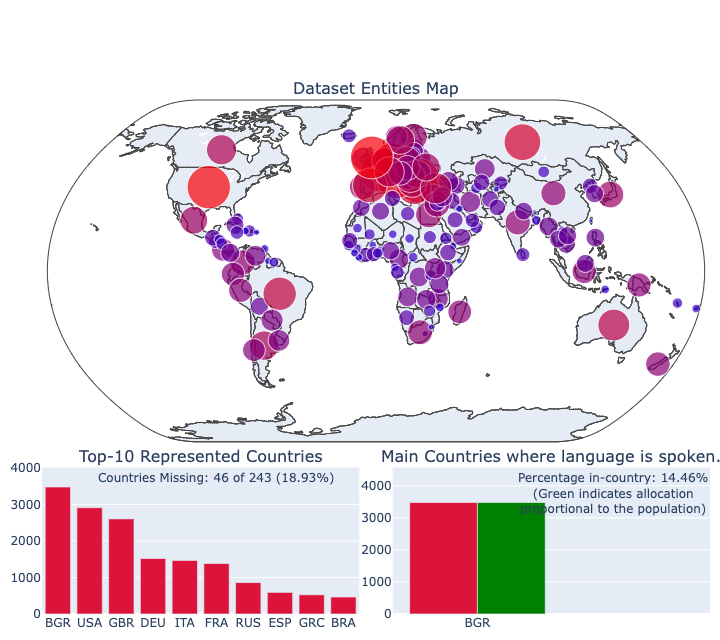}\\
        Azerbaijani & Bulgarian \\
        \includegraphics[width=.45\textwidth]{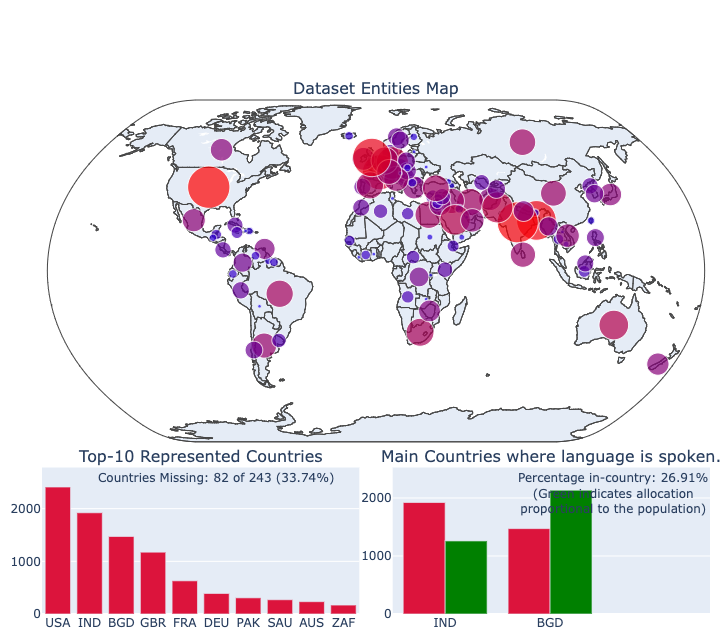} & 
        \includegraphics[width=0.45\textwidth]{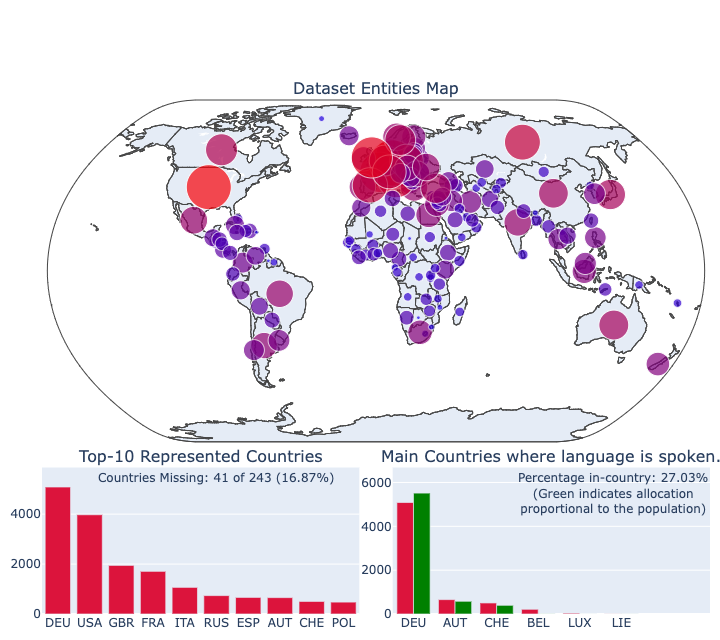}\\
        Bengali & German \\
    \end{tabular}
    \caption{WikiANN Geographic Distributions (Part 1).}
    \label{fig:wikiann1}
\end{figure*}

\begin{figure*}[t]
    \centering
    \begin{tabular}{cc}
    \multicolumn{2}{c}{\textbf{Pan-X (WikiANN) Geographic Coverage}}\\
        \includegraphics[width=.45\textwidth]{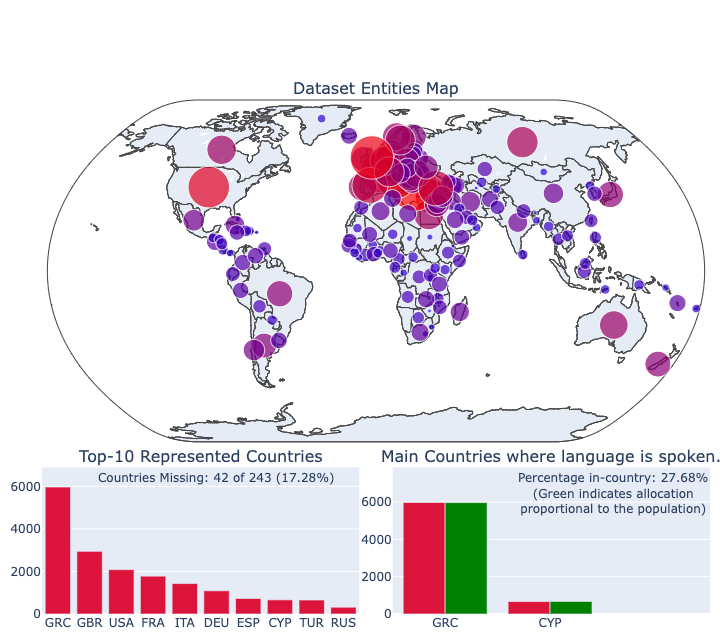} & 
        \includegraphics[width=0.45\textwidth]{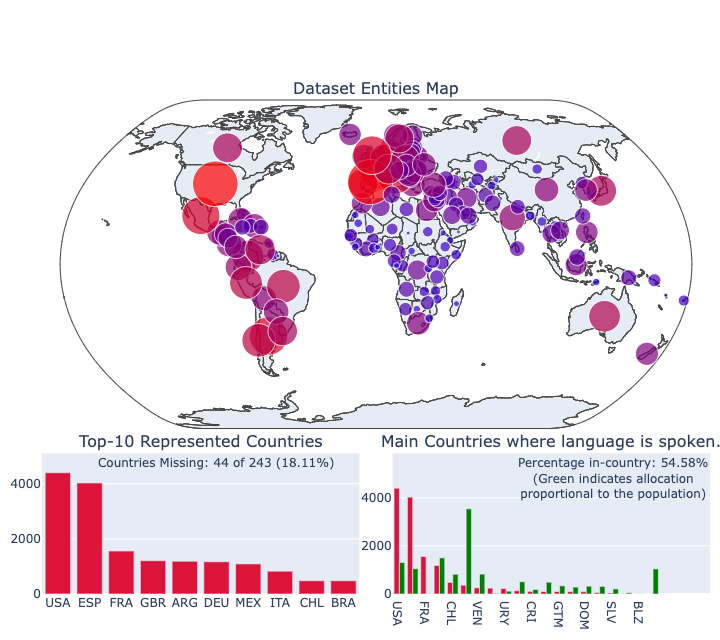}\\
        Greek & Spanish \\
        \includegraphics[width=.45\textwidth]{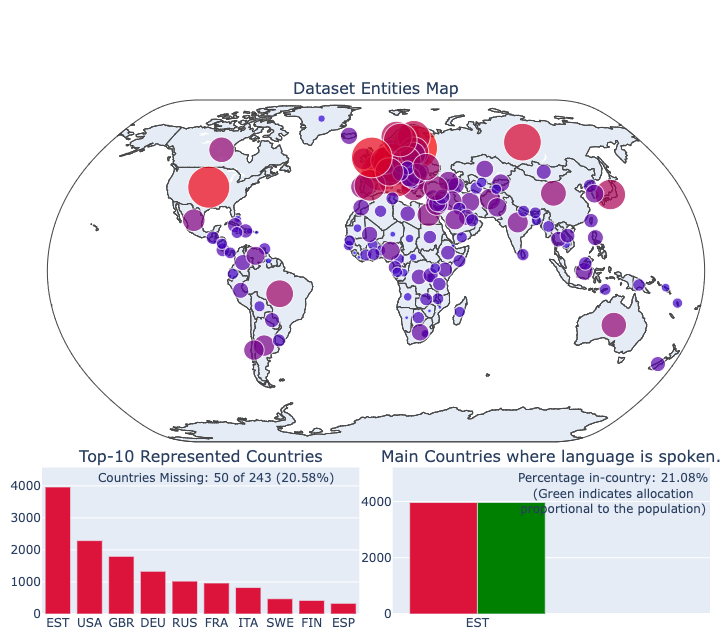} & 
        \includegraphics[width=0.45\textwidth]{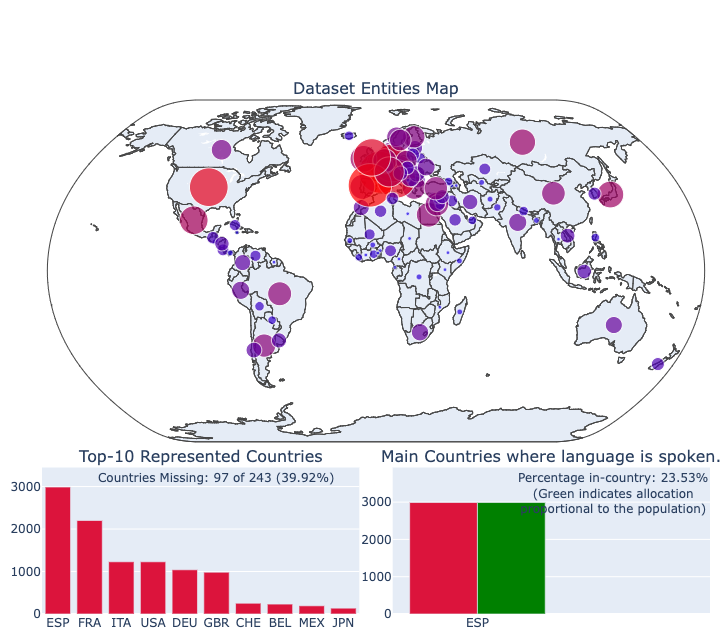}\\
        Estonian & Basque \\
        \includegraphics[width=.45\textwidth]{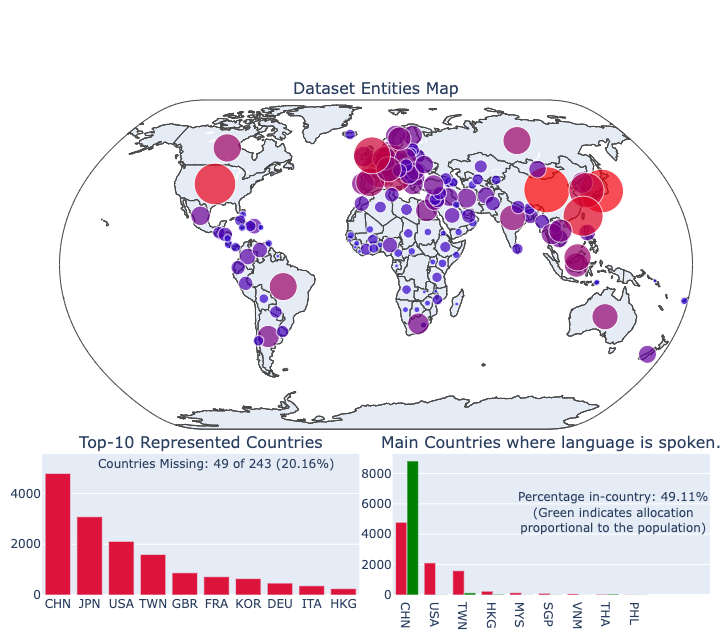} & 
        \includegraphics[width=0.45\textwidth]{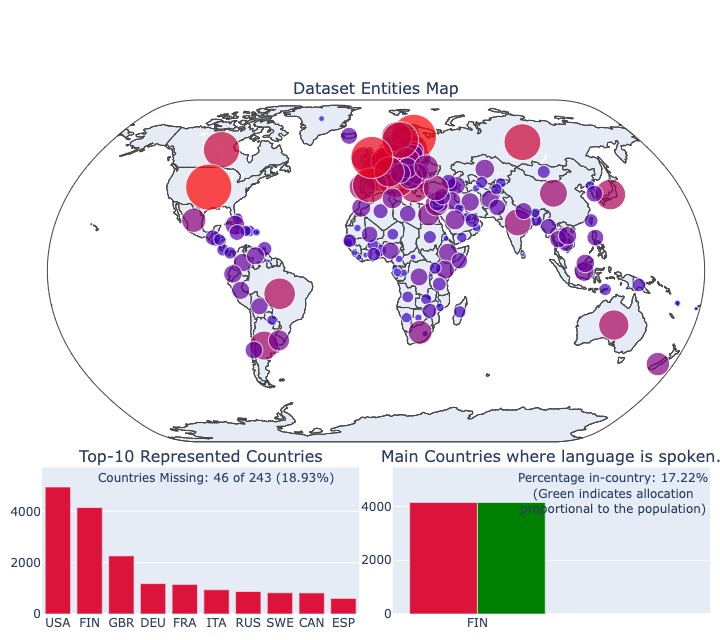}\\
        Chinese & Finnish \\
    \end{tabular}
    \caption{WikiANN Geographic Distributions (Part 2).}
    \label{fig:wikiann2}
\end{figure*}

\begin{figure*}[t]
    \centering
    \begin{tabular}{cc}
    \multicolumn{2}{c}{\textbf{Pan-X (WikiANN) Geographic Coverage}}\\
        \includegraphics[width=.45\textwidth]{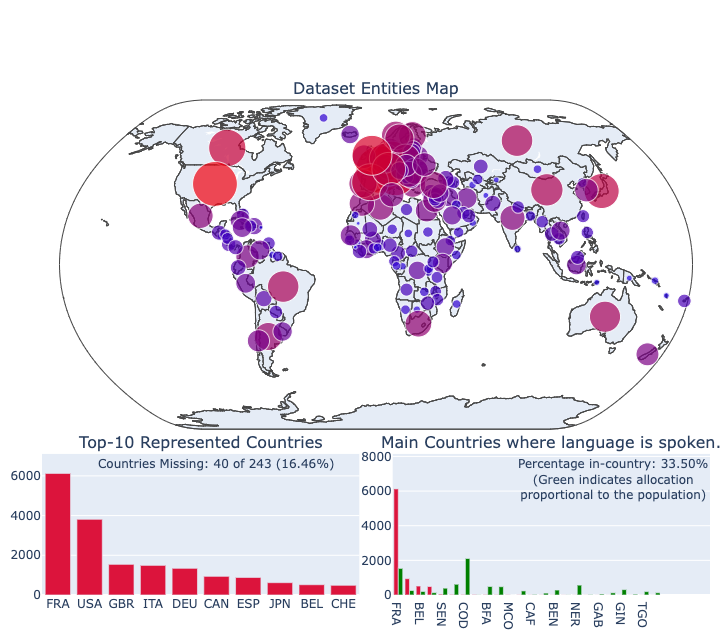} & 
        \includegraphics[width=0.45\textwidth]{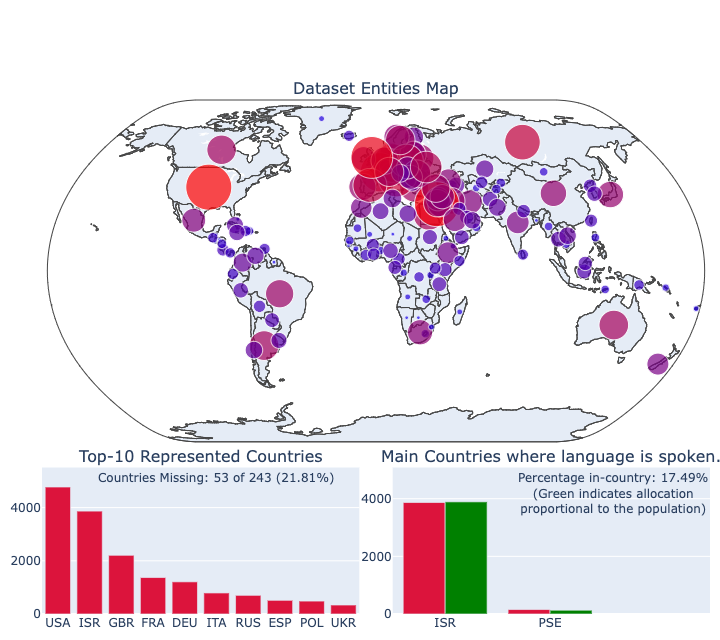}\\
        French & Hebrew \\
        \includegraphics[width=.45\textwidth]{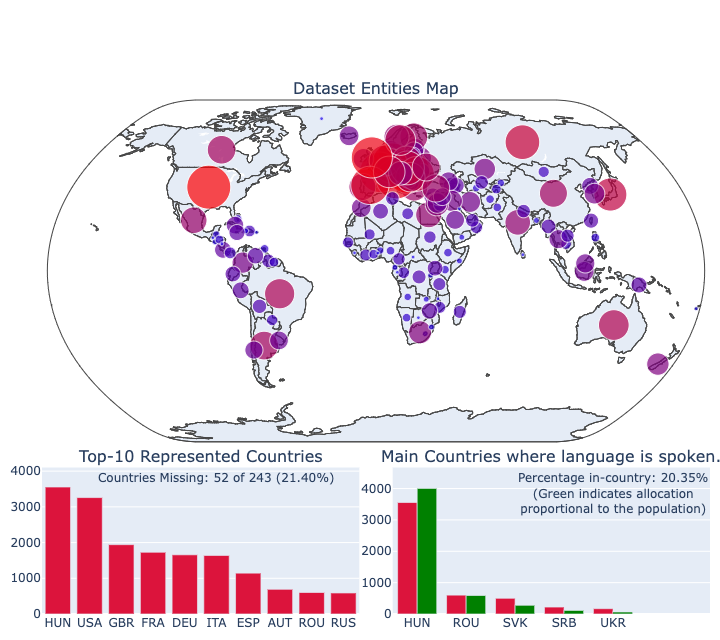} & 
        \includegraphics[width=0.45\textwidth]{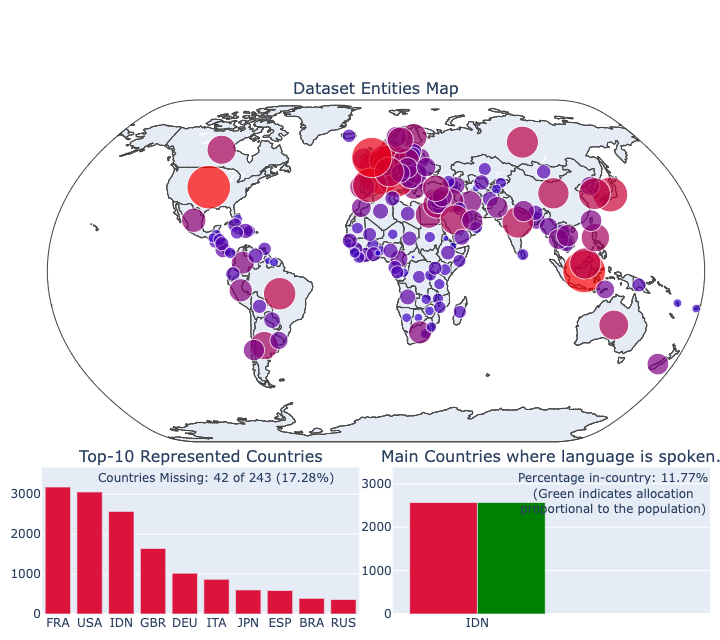}\\
        Hungarian & Indonesian \\
        \includegraphics[width=.45\textwidth]{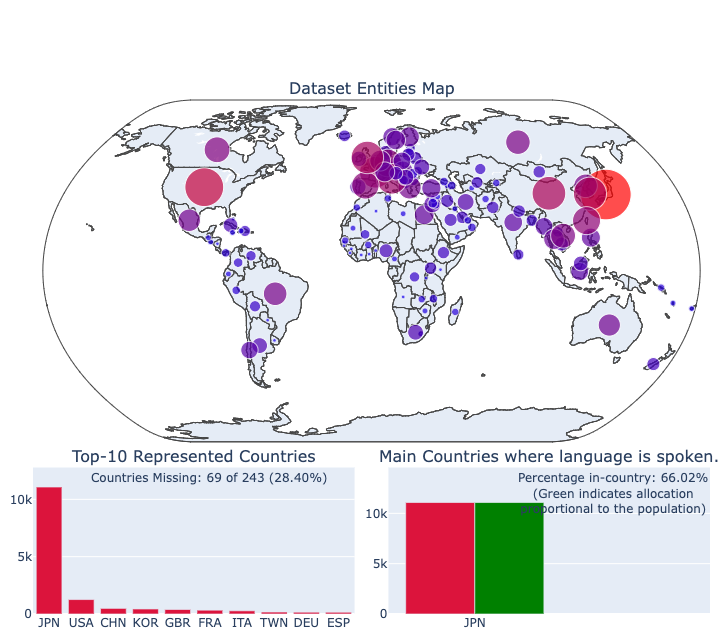} & 
        \includegraphics[width=0.45\textwidth]{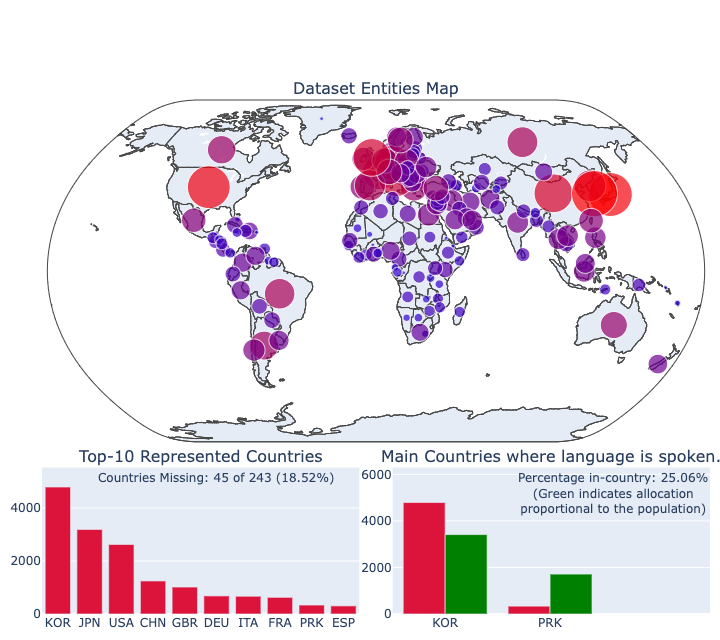}\\
        Japanese & Korean \\
    \end{tabular}
    \caption{WikiANN Geographic Distributions (Part 3).}
    \label{fig:wikiann3}
\end{figure*}

\begin{figure*}[t]
    \centering
    \begin{tabular}{cc}
    \multicolumn{2}{c}{\textbf{Pan-X (WikiANN) Geographic Coverage}}\\
        \includegraphics[width=.45\textwidth]{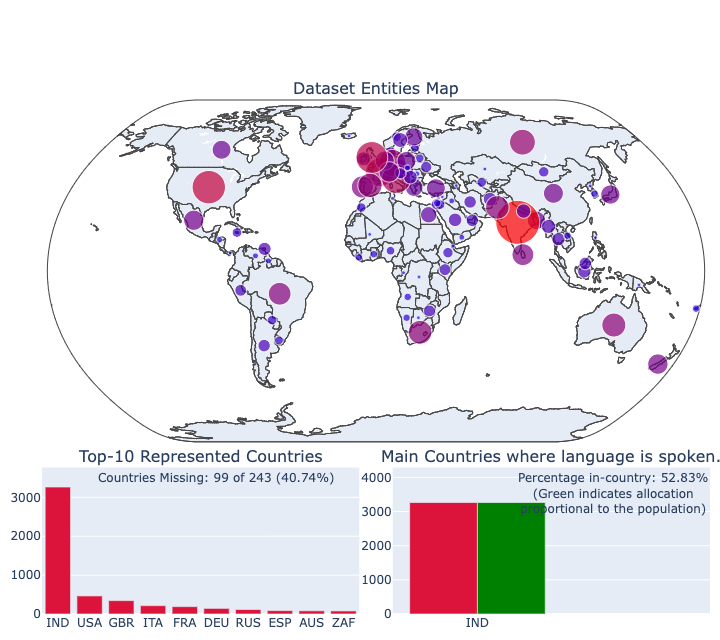} & 
        \includegraphics[width=.45\textwidth]{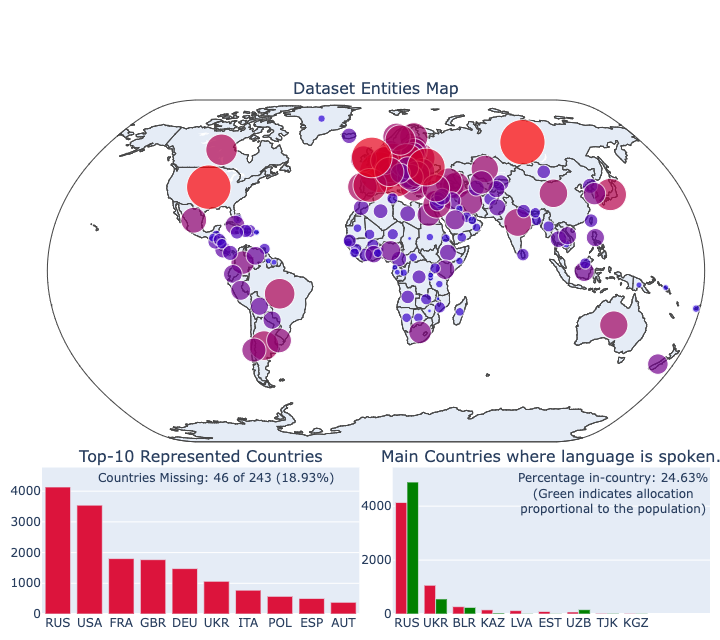}\\
        Marathi & Russian \\
        \includegraphics[width=.45\textwidth]{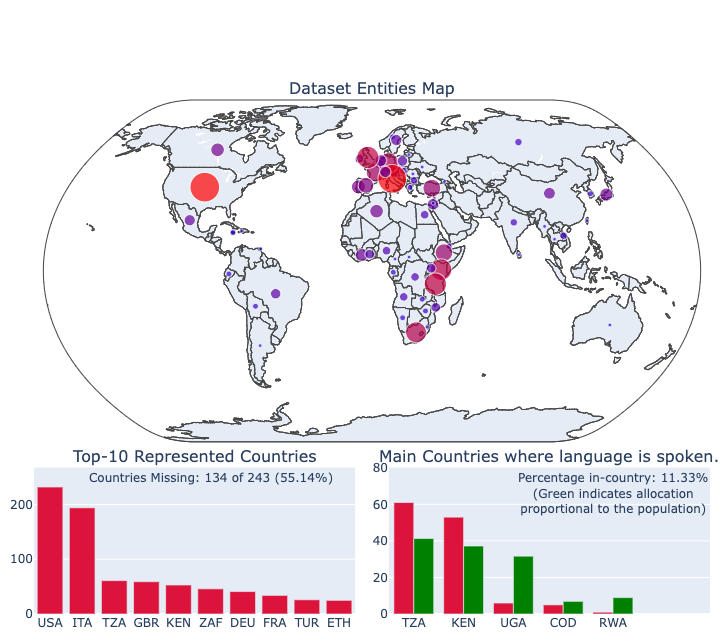} & 
        \includegraphics[width=0.45\textwidth]{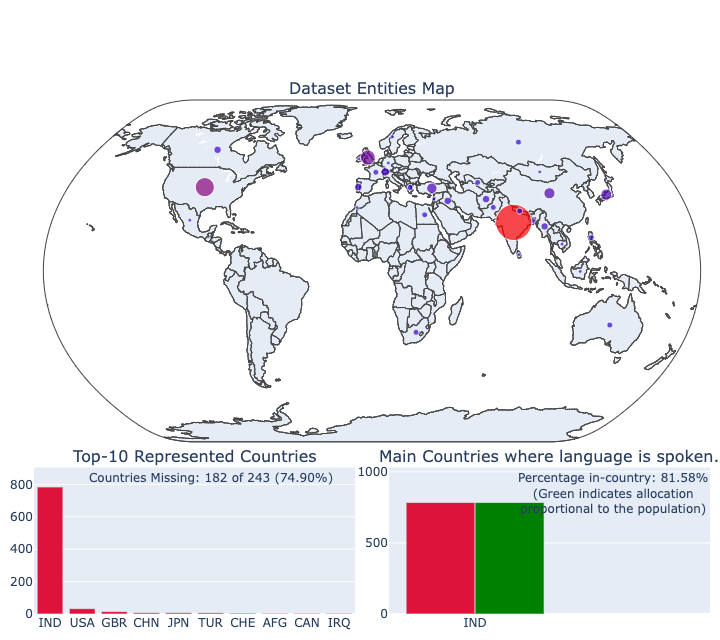}\\
        Swahili & Telegu \\
        \includegraphics[width=.45\textwidth]{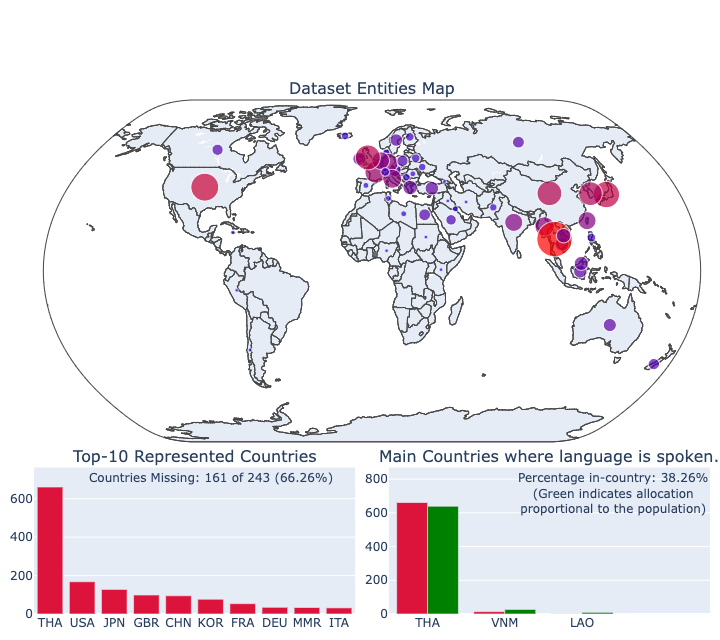} & 
        \includegraphics[width=0.45\textwidth]{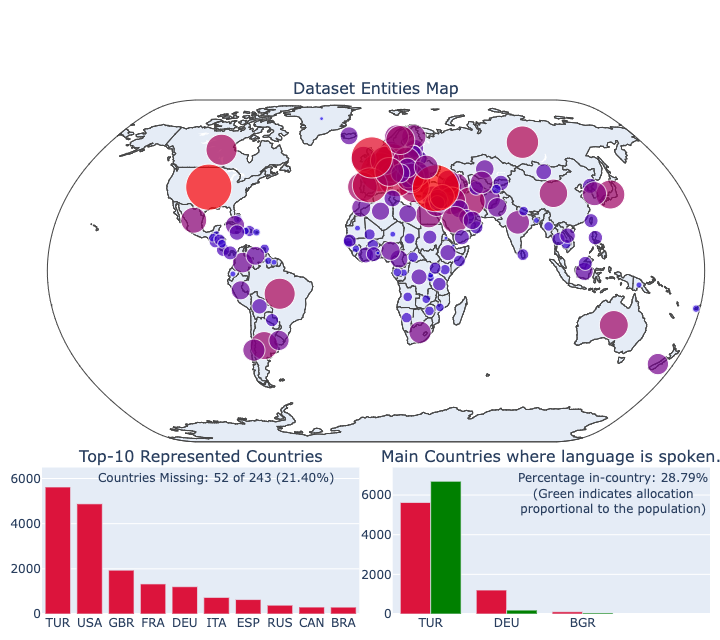}\\
        Thai & Turkish  \\
    \end{tabular}
    \caption{WikiANN Geographic Distributions (Part 4).}
    \label{fig:wikiann4}
\end{figure*}

\begin{figure*}[t]
    \centering
    \begin{tabular}{cc}
    \multicolumn{2}{c}{\textbf{Pan-X (WikiANN) Geographic Coverage}}\\
        \includegraphics[width=.45\textwidth]{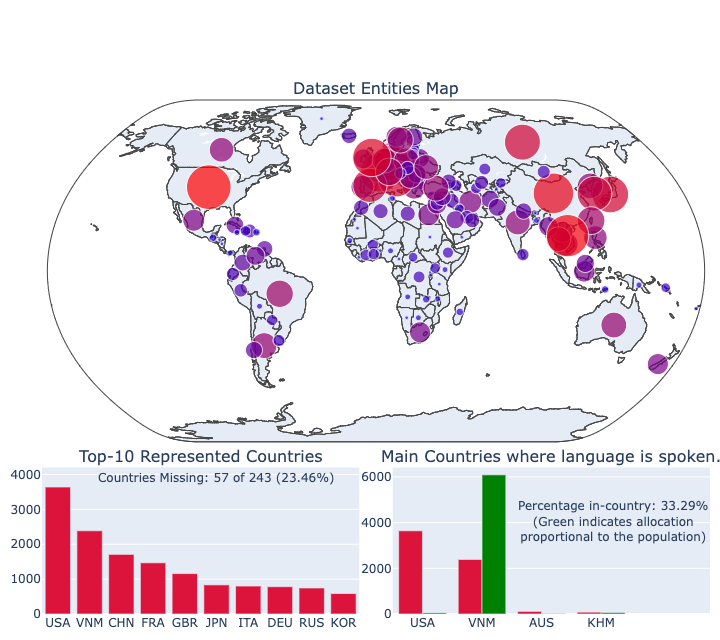} & 
        \includegraphics[width=0.45\textwidth]{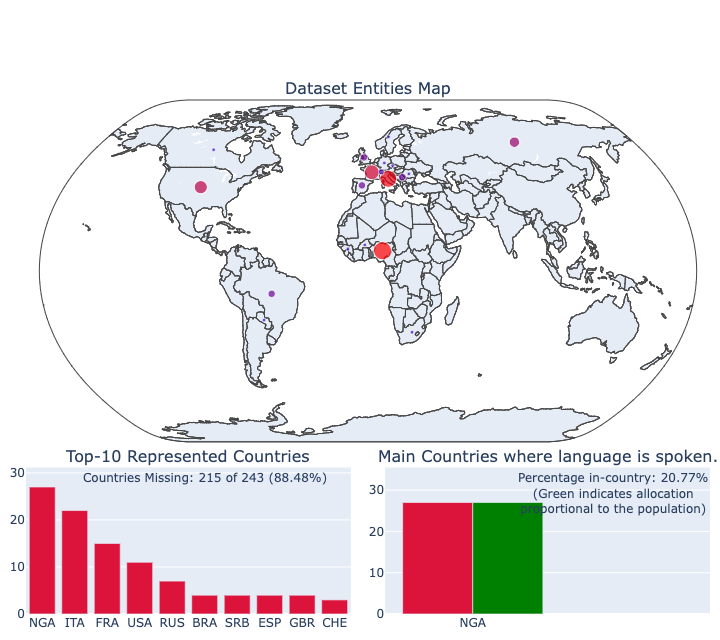}\\
        Vietnamese & Yoruba \\
    \end{tabular}
    \caption{WikiANN Geographic Distributions (Part 5).}
    \label{fig:wikiann5}
\end{figure*}

\begin{figure*}[t]
    \centering
    \begin{tabular}{c}
    \multicolumn{1}{c}{\textbf{SQuAD Geographic Coverage}} \\
        \includegraphics[width=.5\textwidth]{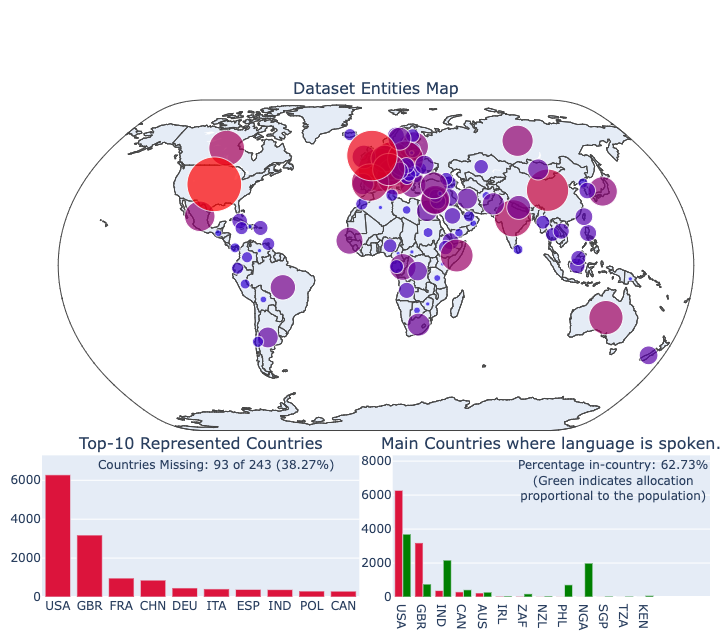} \\
    \end{tabular}
    \caption{SQuAD Geographic Distributions.}
    \label{fig:squad}
\end{figure*}

We present all dataset maps for the datasets we study:
\begin{itemize}
    \item MasakhaNER languages are available in Figures~\ref{fig:masakhaner1} and~\ref{fig:masakhaner2}.
    \item TydiQA languages are available in Figures~\ref{fig:tydiqa1} and~\ref{fig:tydiqa2}.
    \item WikiANN (panx) languages are available in Figures~\ref{fig:wikiann1} through~\ref{fig:wikiann5}.
    \item SQuAD (English) in Figure~\ref{fig:squad}.
\end{itemize}

\section{NER Dataset Socioeconomic Factors}
\label{app:ner_factors}
\begin{table*}[t]
    \centering
    \small
    
    \begin{tabular}{lrr|rr|rr}
    \toprule
        & \multicolumn{2}{c}{X-FACTR (11)} & \multicolumn{2}{c}{MasakhaNER (10)} & \multicolumn{2}{c}{WikiANN (48)} \\
        & \textbf{Explained} & \textbf{} & \textbf{Explained} & \textbf{} & \textbf{Explained} & \textbf{}  \\
        \textbf{Factors $\phi$} & \textbf{Variance} & \textbf{MAE}  & \textbf{Variance} & \textbf{MAE}   & \textbf{Variance} & \textbf{MAE}  \\
    \midrule
         \texttt{pop} & 0.356 & 0.457 & 0.300 & 0.295 & 0.387 & 0.470\\ 
         \texttt{gdp} & 0.516 & 0.407 & 0.341 & 0.295 & 0.575 & 0.382\\
         \texttt{geo} & 0.022 & 0.585 & 0.100 & 0.359 & 0.069 & 0.586\\
    \midrule
        \texttt{pop+gdp} & 0.495 & 0.403 & 0.348 & 0.285 & 0.553 & 0.388\\ 
        \texttt{pop+geo} & 0.356 & 0.455 & 0.369 & 0.290 & 0.399 & 0.467\\ 
        \texttt{geo+gdp} & \textbf{0.521} & \textbf{0.398} & \textbf{0.443} & \textbf{0.284} & \textbf{0.591} & \textbf{0.376}\\ 
    \midrule
        \texttt{pop+gdp+geo} & 0.504 & \textbf{0.398} & 0.440 & 0.285 & 0.572 & 0.380\\ 
    \bottomrule
    \end{tabular}
    \caption{Empirical comparison of factors on NER datasets, averaging over their respective languages (number in parentheses). We report the five-fold cross-validation explained variance and mean absolute error of a linear model.}
    \label{tab:ner_factors}
    \vspace{-1em}
\end{table*}

Table~\ref{tab:factors} presents the same analysis as the one described in Section~\ref{sec:correlates} for the X-FACTR and the NER datasets. The trends are similar to the QA datasets, with GDP being the best predictor and including population statistics hurting the explained variance.

\section{Socioeconomic Correlates Breakdown}
\label{app:correlates}

You can find the breakdown of the socioeconomic correlates in Table~\ref{tab:tydiqa-correlates} for TyDi-QA, Table~\ref{tab:masakhaner-correlates} for MasakhaNER, and Table~\ref{tab:wikiann-correlates} for WikiANN.

\iffinal
\begin{table}[t]
    \centering
    \begin{tabular}{@{}l@{ }l@{ }r@{ }r@{}}
    \toprule
        & & \multicolumn{2}{c}{\texttt{geo+gdp}} \\
        \textbf{Language} & \textbf{Country} & \textbf{Expl. Var.} & \textbf{Mean Error}  \\
    \midrule
        Greek & GRC & 0.586 & 0.343 \\
        Yoruba & NGA & 0.575 & 0.219 \\
        Bengali & BGD & 0.552 & 0.349 \\
        Marathi & IND & 0.587 & 0.29 \\
        French & FRA & 0.569 & 0.452 \\
        Hebrew & ISR & 0.604 & 0.369 \\
        Hungarian & HUN & 0.621 & 0.375 \\
        Russian & RUS & 0.601 & 0.406 \\
        Spanish & ESP & 0.552 & 0.457 \\
        Turkish & TUR & 0.613 & 0.36 \\
        Vietnamese & VNM & 0.521 & 0.398 \\
    \midrule
        Average & &  0.504 & 0.398 \\ 
    \bottomrule
    \end{tabular}
    \caption{Language breakdown of the most predictive factors ($\phi_{\text{geo}}$ and $\phi_{\text{gdp}}$) on X-FACTR dataset.}
    \label{tab:xfactr-correlates}
\end{table}
\fi

\begin{table}[t]
    \centering
    \begin{tabular}{@{}l@{ }l@{ }r@{ }r@{}}
    \toprule
        & & \multicolumn{2}{c}{\texttt{geo+gdp}} \\
        \textbf{Language} & \textbf{Country} & \textbf{Expl. Var.} & \textbf{Mean Error}  \\
    \midrule
Arabic & SAU & 0.501 & 0.415\\
Bengali & BGD & 0.498 & 0.385\\
English & USA & 0.562 & 0.335\\
Finnish & FIN & 0.566 & 0.376\\
Indonesian & IDN & 0.515 & 0.387\\
Japanese & JPN & 0.558 & 0.388\\
Korean & KOR & 0.546 & 0.336\\
Russian & RUS & 0.522 & 0.400\\
Swahili & KEN & 0.428 & 0.469\\
Telugu & IND & 0.534 & 0.294\\
Thai & THA & 0.550 & 0.333\\
    \midrule
        Average & & 0.550 & 0.333  \\ 
    \bottomrule
    \end{tabular}
    \caption{Language breakdown of the most predictive factors ($\phi_{\text{geo}}$ and $\phi_{\text{gdp}}$) on the TyDi-QA dataset.}
    \label{tab:tydiqa-correlates}
\end{table}

\begin{table}[t]
    \centering
    \begin{tabular}{@{}l@{ }l@{ }r@{ }r@{}}
    \toprule
        & & \multicolumn{2}{c}{\texttt{geo+gdp}} \\
        \textbf{Language} & \textbf{Country} & \textbf{Expl. Var.} & \textbf{Mean Error}  \\
    \midrule
        \small Amharic & ETH & 0.131 & 0.220 \\
        \small Yoruba & NGA & 0.338 & 0.258\\
        \small Hausa & NGA & 0.321 & 0.317\\
        \small Igbo & NGA & 0.326 & 0.207\\
        \small Kinyarwanda & RWA & 0.198 & 0.229\\
        \small Luganda & UGA & 0.302 & 0.195\\
        \small Luo & ETH & 0.000 & 0.110\\
        \small Nigerian English & NGA & 0.493 & 0.231\\
        \small Wolof & CMR & 0.378 & 0.160\\
        \small Swahili & KEN & 0.443 & -0.285 \\
    \midrule
        Average & &  0.378 & 0.160 \\ 
    \bottomrule
    \end{tabular}
    \caption{Language breakdown of the most predictive factors ($\phi_{\text{geo}}$ and $\phi_{\text{gdp}}$) on MasakhaNER dataset.}
    \label{tab:masakhaner-correlates}
\end{table}

\begin{table}[t]
    \centering
    \begin{tabular}{@{}l@{ }l@{ }r@{ }r@{}}
    \toprule
        & & \multicolumn{2}{c}{\texttt{geo+gdp}} \\
        \textbf{Language} & \textbf{Country} & \textbf{Expl. Var.} & \textbf{Mean Error}  \\
    \midrule
        af & ZAF & 0.497 & 0.338 \\
ar & SAU & 0.570 & 0.454 \\
az & AZE & 0.566 & 0.395 \\
bg & BGR & 0.511 & 0.475 \\
bn & BGD & 0.442 & 0.502 \\
de & DEU & 0.613 & 0.402 \\
el & GRC & 0.484 & 0.456 \\
es & ESP & 0.497 & 0.462 \\
et & EST & 0.565 & 0.398 \\ 
eu & ESP & 0.565 & 0.387 \\
fa & IRN & 0.589 & 0.426\\
fi & FIN & 0.590 & 0.411\\
fr & FRA & 0.597 & 0.408\\
gu & IND & 0.068 & 0.030\\
he & ISR & 0.551 & 0.456\\
hi & IND & 0.529 & 0.279\\
hu & HUN & 0.563 & 0.451\\
id & IDN & 0.488 & 0.442\\
it & ITA & 0.569 & 0.436\\
ja & IDN & 0.591 & 0.343\\
jv & JPN & 0.062 & 0.069\\
ka & GEO & 0.474 & 0.435\\
kk & KAZ & 0.411 & 0.205\\
ko & KOR & 0.519 & 0.423\\
lt & LTU & 0.533 & 0.395\\
ml & IND & 0.495 & 0.367\\
mr & IND & 0.530 & 0.320\\
ms & MYS & 0.496 & 0.463\\
my & MMR & 0.105 & 0.038\\
nl & NLD & 0.582 & 0.435\\
pa & IND & 0.052 & 0.064\\
pl & POL & 0.584 & 0.436\\
pt & PRT & 0.567 & 0.432\\
qu & PER & 0.301 & 0.090\\
ro & ROU & 0.581 & 0.436\\
ru & RUS & 0.576 & 0.435\\
sw & KEN & 0.402 & 0.223\\
ta & LKA & 0.524 & 0.367\\
te & IND & 0.351 & 0.107\\
th & THA & 0.567 & 0.215\\
tl & PHL & 0.473 & 0.399\\
tr & TUR & 0.619 & 0.409\\
uk & UKR & 0.576 & 0.447\\
ur & PAK & 0.512 & 0.463\\
vi & VNM & 0.557 & 0.440\\
yo & NGA & 0.079 & 0.086\\
zh & CHN & 0.591 & 0.376\\
    \midrule
        Average & & 0.591 & 0.376  \\ 
    \bottomrule
    \end{tabular}
    \caption{Language breakdown of the most predictive factors ($\phi_{\text{geo}}$ and $\phi_{\text{gdp}}$) on the WikiANN dataset.}
    \label{tab:wikiann-correlates}
\end{table}

\section{NER Models Confusion Matrices}
\label{app:ner_conf_matrix}
\begin{figure*}
    \centering
    \begin{tabular}{m{.4cm}@{}m{.2cm}m{4cm}m{4cm}m{4cm}}
        && \multicolumn{1}{c}{\textbf{Greek}} & \multicolumn{1}{c}{\textbf{Italian}} & \multicolumn{1}{c}{\textbf{Chinese}} \\
        && \multicolumn{1}{c}{diagonal: 46.0\%} & \multicolumn{1}{c}{diagonal: 59.1\%} & \multicolumn{1}{c}{diagonal: 16.8\%}\\
        \multicolumn{2}{c}{\rotatebox[origin=c]{90}{mBERT}} &
        \includegraphics[scale=0.28]{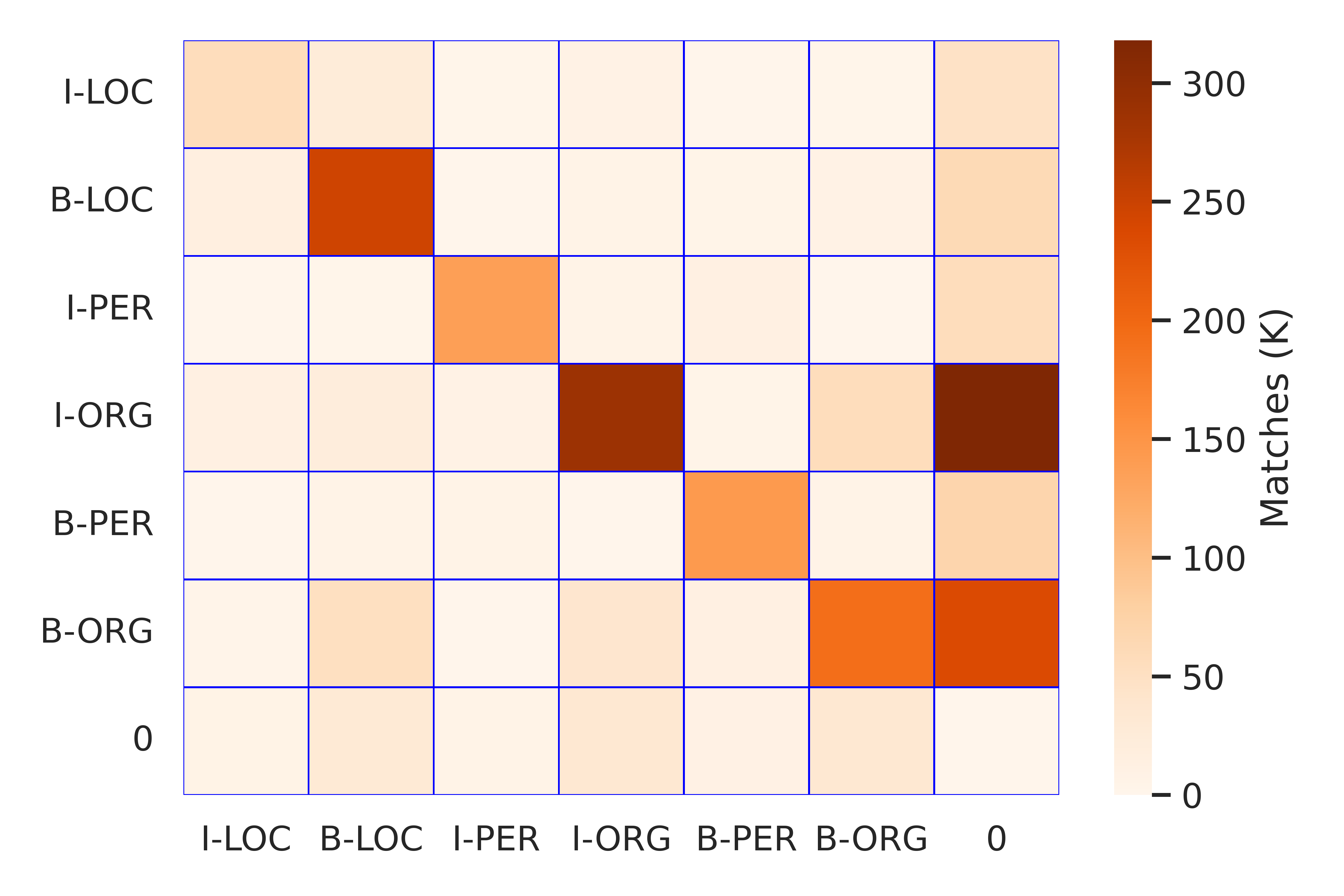} & 
         \includegraphics[scale=0.28]{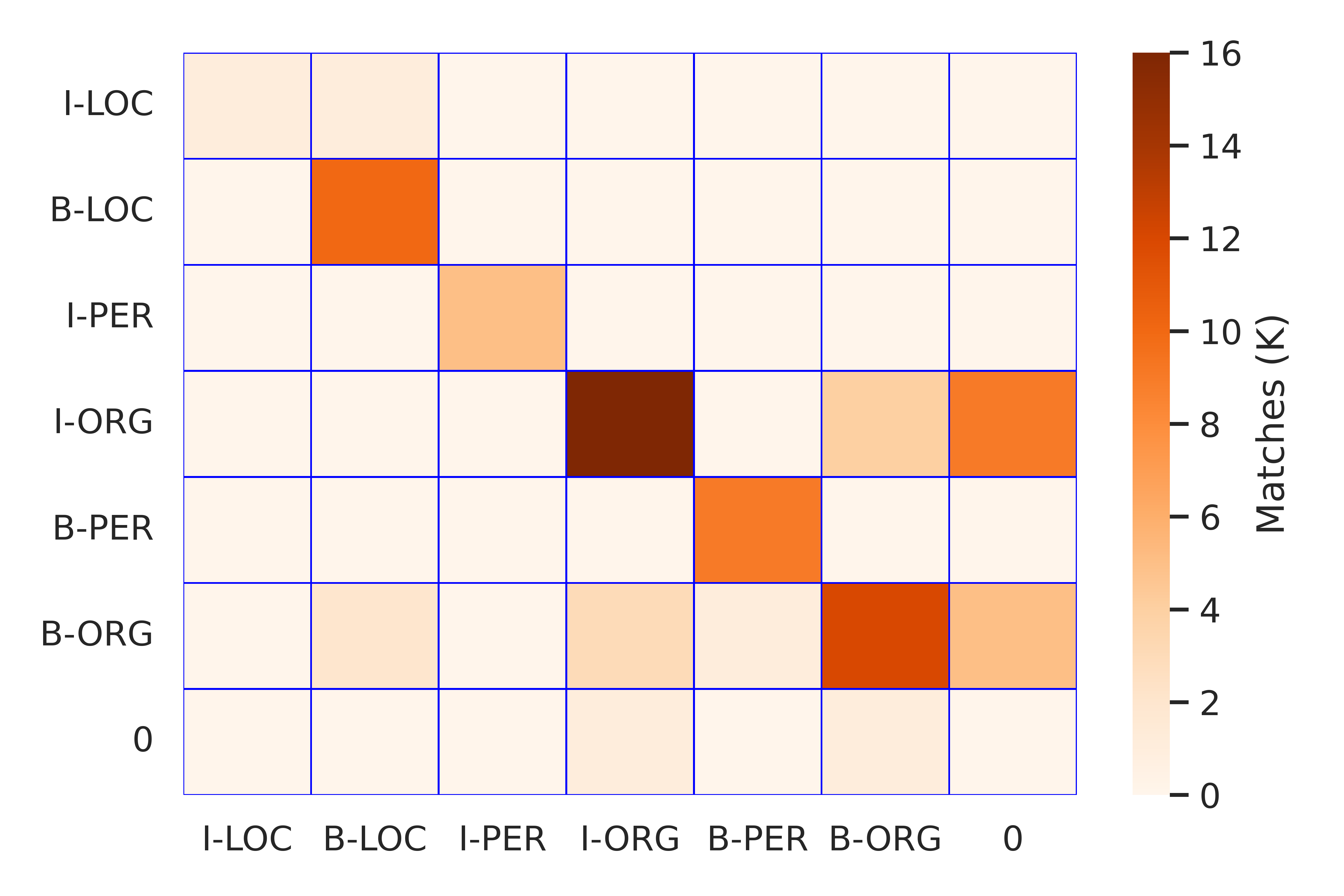} &
        \includegraphics[scale=0.28]{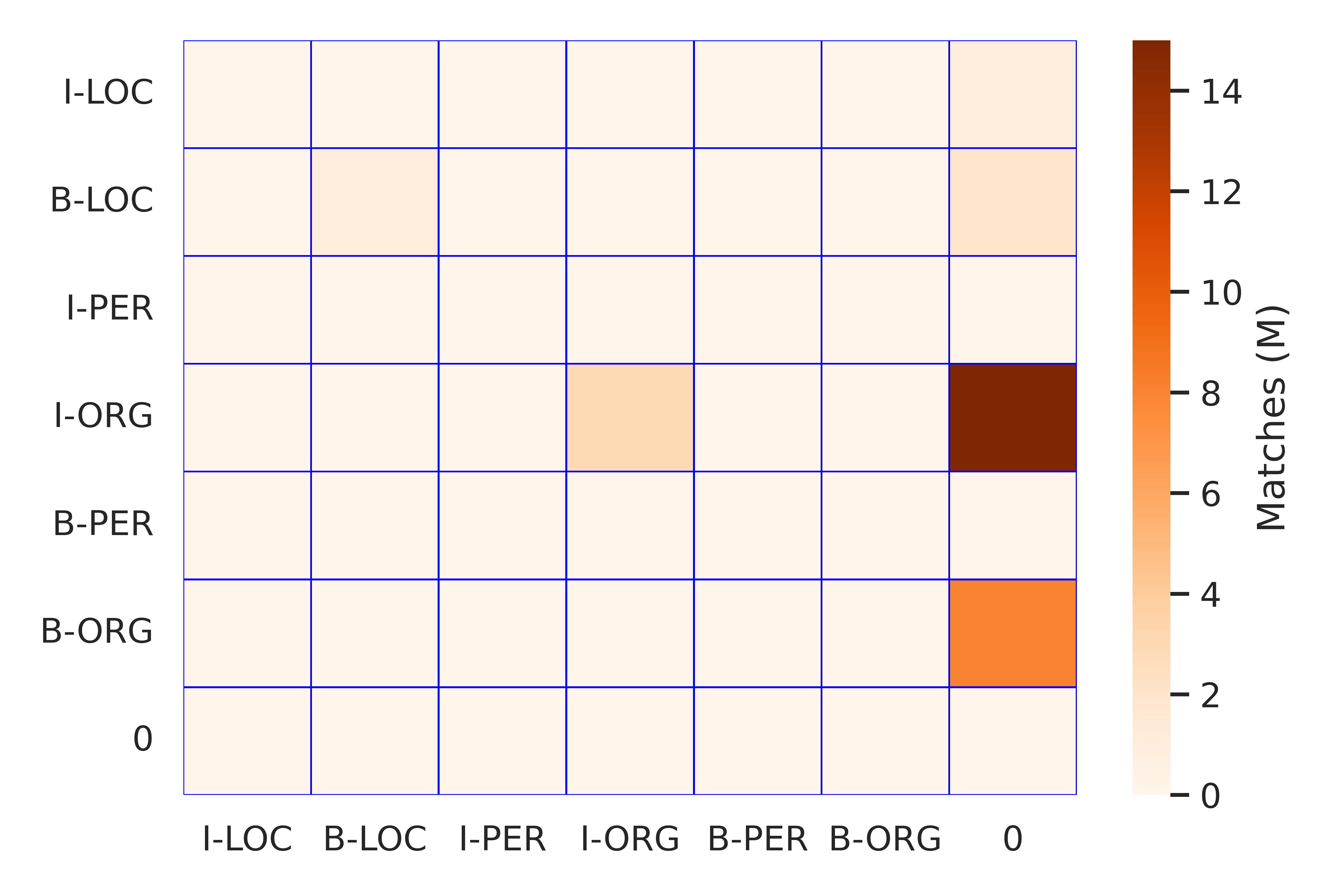} \\
        \iffinal
        && \multicolumn{1}{c}{diagonal: 45.4\%} & \multicolumn{1}{c}{diagonal: 60.5\%} & \multicolumn{1}{c}{diagonal: 11.5\%} \\
         \rotatebox[origin=c]{90}{mBERT} &
         \rotatebox[origin=c]{90}{+aligned} &
        \includegraphics[scale=0.28]{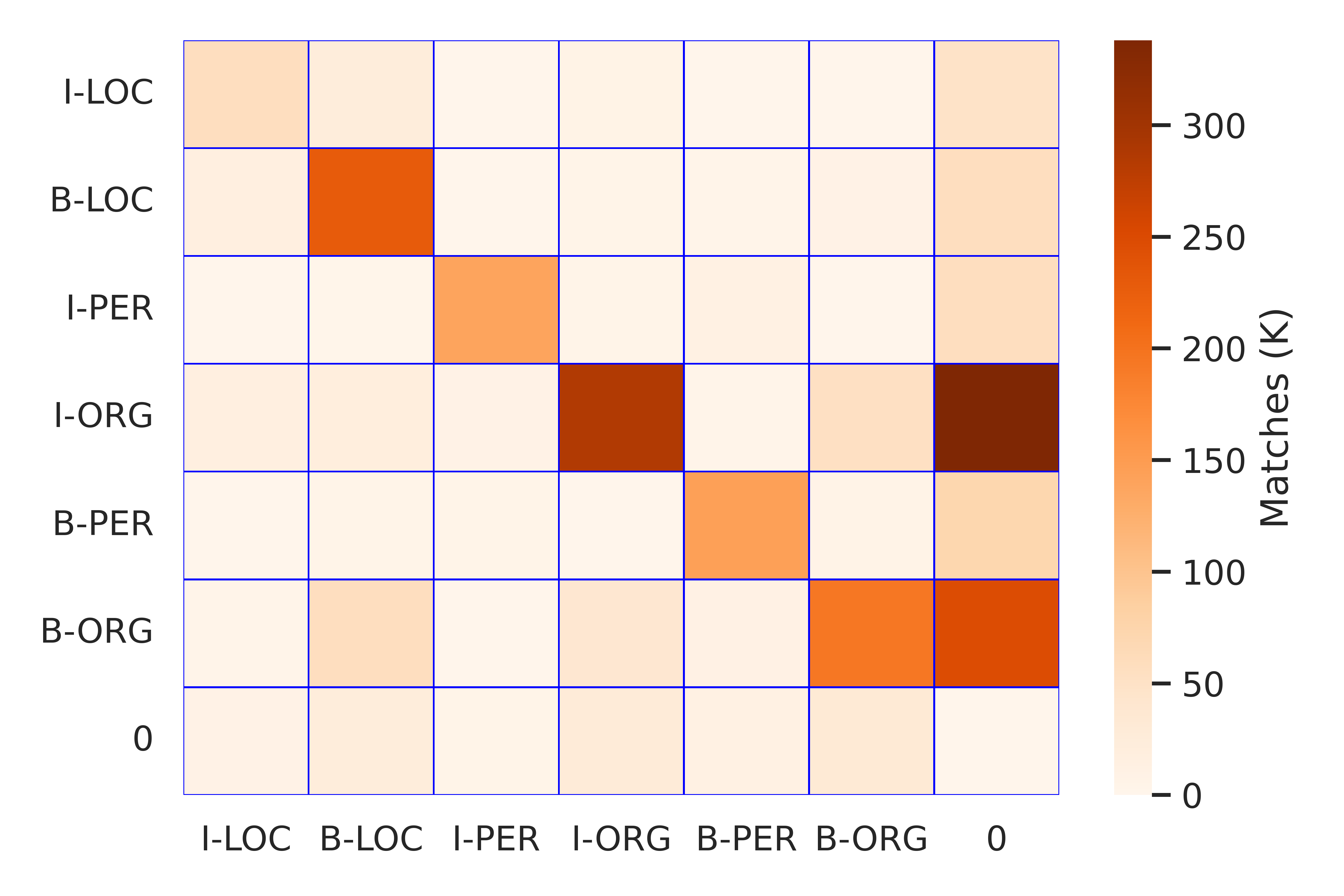} &
         \includegraphics[scale=0.28]{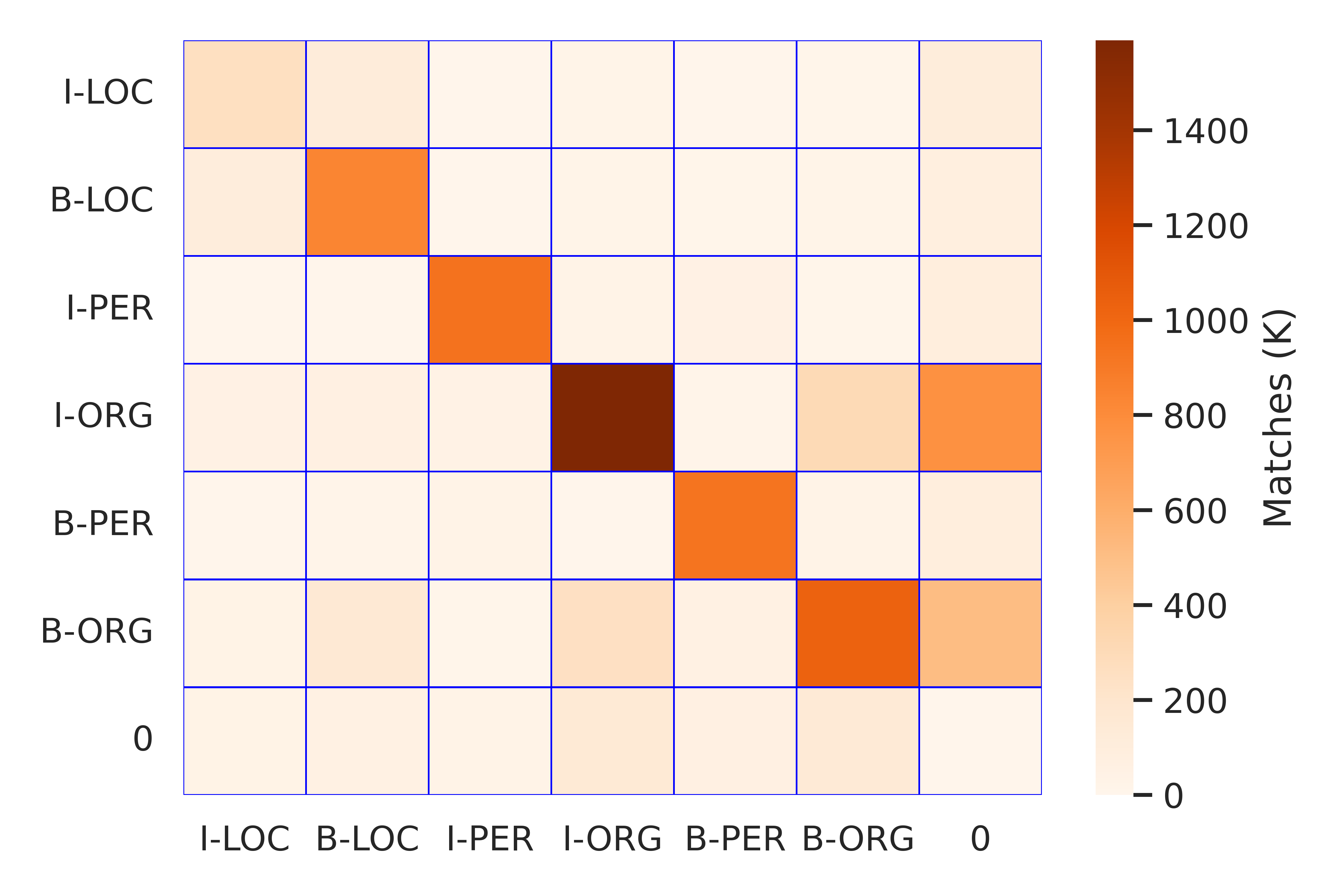}&
        \includegraphics[scale=0.28]{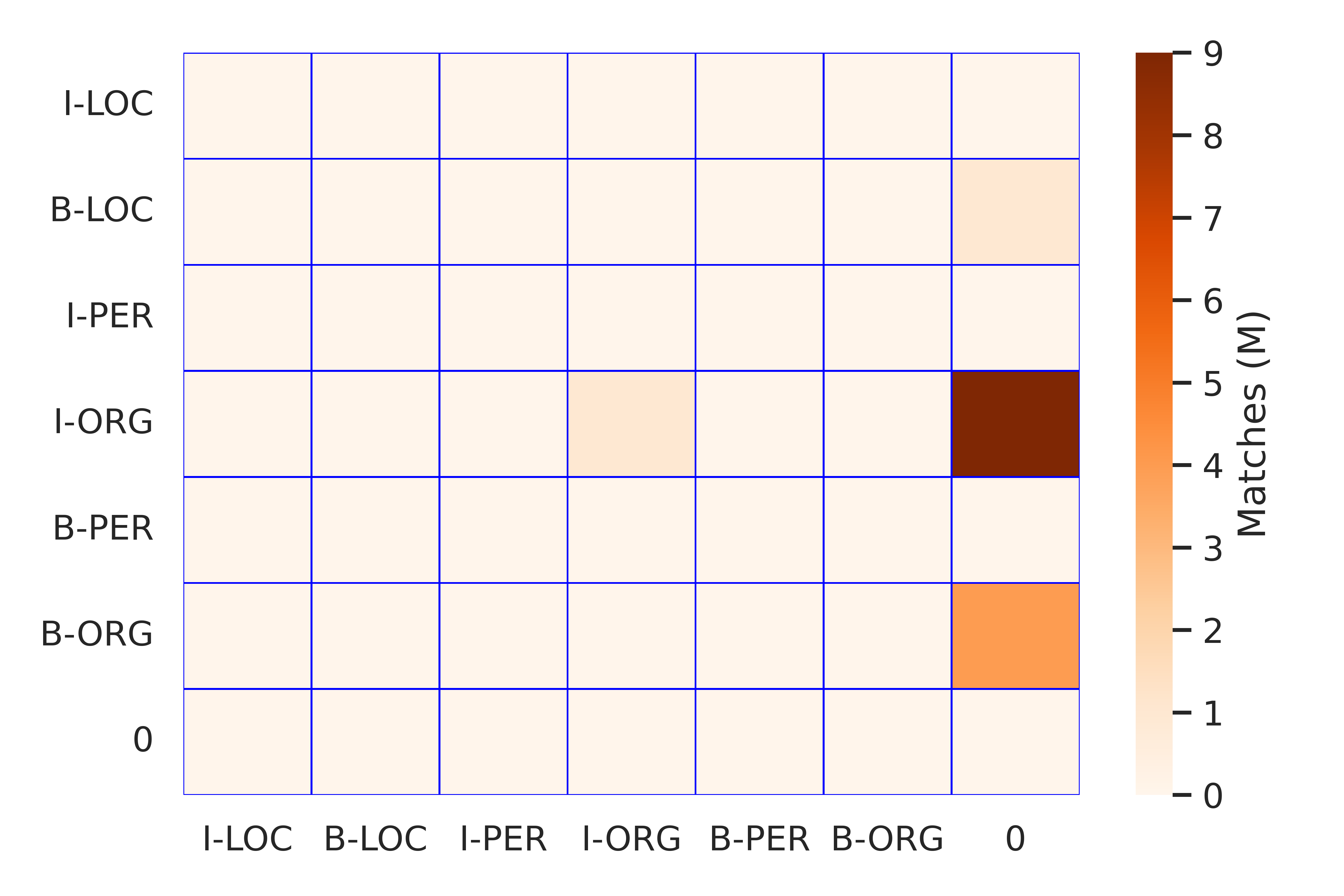}\\
         \fi
    \end{tabular}
    \caption{Confusion matrices for Greek, Italian and Chinese.}
    \label{fig:ner_confusion}
\end{figure*}

See Figure~\ref{fig:ner_confusion} for the confusion matrices of the SpaCy and our WikiANN neural model.

\section{Greek-English NER Error Discussion}
\label{app:ner_discussion}

We find that the mistakes we identify vary significantly by label. In about 75\% of the \texttt{0-LOC} cases it was the Greek-side labels that were wrong in tagging a span as a location. A common pattern we identified (about 35\% of these cases) was the Greek model tagging as location what was actually a month. For instance, in the sentence \texttt{Ton Máio tu 1990 episkéftikan yia tésseris iméres tin Ouggaria}
(\textit{In May 1990 , they visited Hungary for four days.}) the model tags the first two words (``in May") as a location, while the English one correctly leaves them unlabelled.

In the case of \texttt{LOC-0} cases, we found an even split between the English- and the Greek-side labels being wrong (with about 40\% of the sentences each).  Common patterns of mistakes in the English side include tagging persons as locations (e.g. ``Heath" in ``Heath asked the British to heat only one room in their houses over the winter." where ``Heath" corresponds to Ted Heath, a British politician), as well as tagging adjectives, often locative, as locations, such as ``palaeotropical" in  ``Palaeotropical refers to geographical occurrence." and  ``French" in ``A further link [..] by vast French investments and loans [...]". 

Last, in the case of \texttt{0-PER} cases we studied, we found that 62\% of the errors were on the English side. A common pattern was the English-side model not tagging persons when they are the very first token in a sentence, i.e. the first tokens in ``Olga and her husband were left at Ay-Todor.", in ``Friedman once said,  `If you want to see capitalism in action, go to Hong Kong.' ", and in ``Evans was a political activist before [...]" were all tagged as \texttt{0}. To a lesser extent, we observed a similar issue when the person's name followed punctuation, e.g. ``Yavlinsky" in the sentence ``In March 2017 , Yavlinsky stated that he will [...]".

\section{Comparing X-FACTR to mLAMA} These two similar projects aim at testing the memorization abilities of large language models (X-FACTR and multi-LAMA~\cite[mLAMA;][]{kassner-etal-2021-multilingual}) -- see corresponding Figures in Table~\ref{fig:xfactr}. Both of these build on top of Wikidata and the mTREx dataset. Hence, their English portions are equally representative of English speakers, sufferring from under-representation of English speakers of the Global South. For the other language, however, mLAMA translates English prompts and uses entity-relation triples mined from the English portion of Wikidata, unlike X-FACTR which uses different data for each language, mined from their respective portion of Wikidata. Both are still western-biased, since they rely on Wikipedia, but one (X-FACTR) is better at giving an indication of potential downstream utility to users.

\end{document}